\documentclass[preprint,5p,times,twocolumn]{elsarticle}

\usepackage{amssymb}
\usepackage{amsmath}
\usepackage{graphicx}
\usepackage{stfloats}
\usepackage{amsmath, amsthm}
\usepackage{hyperref}
\usepackage{colortbl}
\usepackage{xcolor}
\usepackage{multirow}
\usepackage{booktabs}
\usepackage{caption}
\usepackage{makecell}
\usepackage{tabularx}

\theoremstyle{definition}
\definecolor{rankone}{RGB}{185,220,95} % green
\definecolor{ranktwo}{RGB}{255,194,73} % orange
\definecolor{rankthree}{RGB}{255,255,0} % yellow

\journal{Nuclear Physics B}

\begin{document}

\begin{frontmatter}



\title{Benchmarking graph-based models for in-silico toxicity prediction in drug discovery} %% Article title


\author[citius]{Noel Suárez-Barro} %% Author name
\author[citius]{Juan Carlos Vidal Aguiar}
\author[citius]{Manuel Lama Penín}

\affiliation[citius]{organization={Centro Singular de Investigación en Tecnoloxías Intelixentes (CiTIUS)},%Department and Organization
            addressline={Universidade de Santiago de Compostela}, 
            city={15705 Santiago de Compostela},
            %postcode={15705}, 
            %state={Galicia},
            country={Spain}}

%% Abstract
\begin{abstract}
Drug discovery is a costly and high-risk process, where toxicity-related failures remain a major cause of attrition in both preclinical and clinical stages. As a result, accurate early prediction of chemical toxicity is essential to reduce downstream costs and improve compound prioritization. In this context, graph deep learning (GDL) has emerged as a powerful paradigm for toxicity prediction, leveraging molecular graph representations to learn directly from chemical structure with improved expressivity over traditional approaches.

Despite the growing number of proposed models, current literature-based comparisons are often difficult to interpret due to inconsistencies in datasets, preprocessing pipelines, and evaluation protocols. To address this limitation, we introduce a unified and standardized benchmarking framework for GDL-based toxicity prediction. We systematically evaluate more than 20 representative approaches under consistent experimental conditions and across multiple datasets and partitioning strategies, enabling a fair and reproducible comparison of model performance. In addition, we complement this empirical study with a structured literature analysis to contextualize existing methodological trends and performance claims. Our results provide a clearer and more reliable assessment of the current state of the field, highlighting both the strengths and limitations of existing graph-based approaches. To support transparency and reproducibility, we release our benchmarking framework as open-source software \url{https://gitlab.citius.gal/noel.suarez/benchtox}, allowing the community to evaluate and compare models under consistent conditions.
\end{abstract}

\raggedbottom

\begin{graphicalabstract}
\begin{figure*}[t]
\centering
\includegraphics[width=\textwidth]{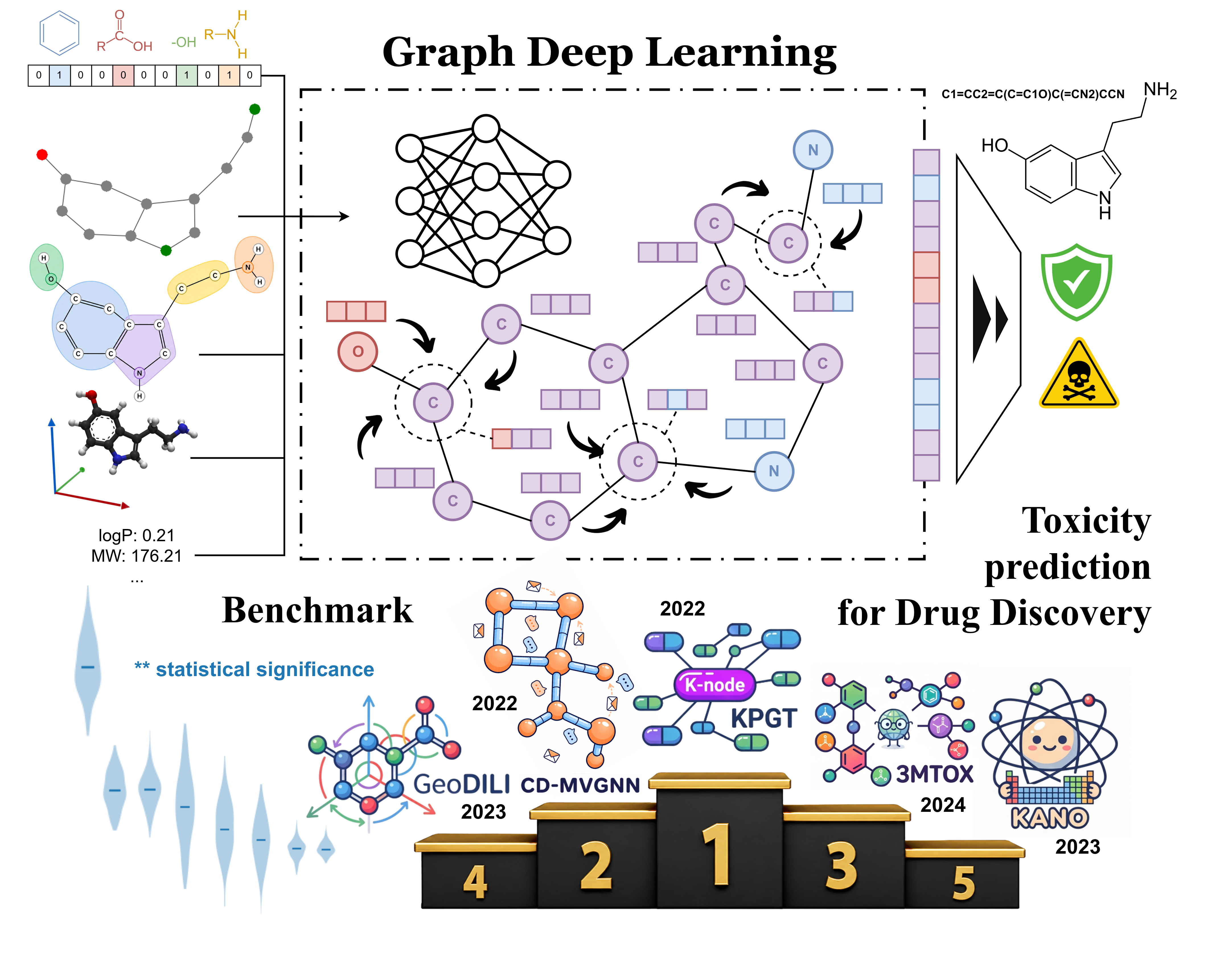}
\end{figure*}
\end{graphicalabstract}

\begin{highlights}
\item A unified benchmarking framework for graph-based toxicity prediction.
\item Systematic evaluation of 20+ models under consistent experimental conditions.
\item Fair and reproducible comparison across multiple datasets and partitioning strategies.
\item Analysis of methodological trends and performance claims in the literature.
\item Open-source release of the benchmarking framework to support community research.
\end{highlights}


\end{frontmatter}



\section{Introduction} %% CHAPTER 1

%% Drug discovery
Development of new pharmaceuticals —usually referred as \textit{Drug Discovery}— remains a lengthy, costly, and high-risk process in which safety and efficacy must be carefully balanced. Despite continuous technological advances, bringing a drug from initial identification to clinical approval can take over a decade and require investments of several billion dollars \cite{principles_of_early_drug_discovery}. A significant portion of these costs is driven by late-stage failures, many of which are associated with toxicity issues \cite{rev_VanTran}. As a result, early and reliable toxicity assessment has become a central objective in modern drug development, aiming to identify unsafe compounds as soon as possible and thereby reduce downstream attrition.

More broadly, drug discovery encompasses a spectrum of computational tasks that support molecular design and optimization, including target identification, prediction of drug–target interactions and binding affinities, de novo molecule generation, and estimation of key physicochemical and bioactivity properties such as solubility and permeability \cite{Zhang2025Computational}. Within this landscape, toxicity prediction plays a critical role in prioritizing safer candidates and minimizing reliance on costly experimental assays and animal testing \cite{Lee2025Recent}.

%% Toxicity
Toxicity itself is a complex, multifactorial phenomenon arising from the interplay between molecular structure, physicochemical properties, and biological processes across multiple scales. Conventional approaches, including \textit{in vitro} and \textit{in vivo} assays, remain essential but are often limited by high costs, long turnaround times, and ethical concerns. Consequently, computational methods —particularly quantitative structure–activity relationship (QSAR) models— have been widely adopted to complement experimental studies by linking molecular descriptors to toxicological outcomes \cite{QSAR_comprehensive}. However, traditional QSAR approaches frequently struggle to generalize across diverse chemical spaces and to capture the nonlinear and context-dependent nature of biological responses \cite{QSAR_Elsevier}. Toxic effects can vary substantially depending on factors such as dose, route of administration, metabolism, and inter-individual or inter-species variability, complicating both label definition and model transferability.

In practice, toxicity prediction for small molecules is typically framed as a supervised learning problem, where models take molecular representations as input and predict toxicity indicators derived from experimental assays, ranging from binary safety labels to continuous dose–response measures. The ultimate goal is to generalize this knowledge to unseen compounds and improve the early identification of adverse effects, which often stem from subtle structural features, reactive metabolites, or unintended off-target interactions \cite{AIM_isMLbetter}.

%% Graph deep learning
With the emergence of large-scale omics data and advances in Artificial Intelligence, deep learning methods have redefined the landscape of computational toxicology \cite{QSAR_Oxford, AIM_Multiview_hetgraph}. Among these, graph deep learning (GDL) has emerged as a transformative paradigm \cite{AIM_heteroGNN, AIM_knowledgegraph_DTI, AIM_VAEGANMDA}, leveraging graph-based molecular representations that inherently encode the topology and connectivity of atoms and bonds. Graph neural networks (GNNs) learn data-driven molecular embeddings capable of integrating structural, physicochemical, and contextual biological information, enabling more expressive and generalizable predictive models. Unlike traditional fixed molecular descriptors like fingerprints (FPs), these architectures can automatically infer hierarchical structure–property relationships directly from raw molecular graphs \cite{AIM_QSAR}.

The application of GDL to toxicity prediction offers unprecedented opportunities to enhance accuracy, interpretability, and transferability across datasets and endpoints. Recent advances have demonstrated its potential to uncover subtle molecular patterns underlying adverse effects and mechanistic toxicities. This work provides both a comprehensive overview of graph-based deep learning methods for toxicity prediction and an extensive comparative evaluation of state-of-the-art approaches. Through systematic and reproducible benchmarking under equal experimental conditions of the most influential and widely adopted GDL architectures across multiple public toxicity datasets, we assess their predictive performance, robustness, and generalization capabilities. By combining a critical synthesis of current research with rigorous empirical analysis, this study delineates both the progress achieved and the limitations that define the current landscape of AI-driven drug safety prediction, while highlighting promising directions for future research in computational toxicology.

The main contributions of this work are summarized below:
\begin{itemize}
    \item We provide a unified synthesis of existing approaches and architectures, formalizing them within a \textbf{common framework} that defines the methodological landscape for developing deep learning models in toxicity prediction, as well as the types of information typically employed.
    
    \item We conduct a comprehensive analysis and \textbf{systematic classification} of the reviewed methods according to the criteria established within this framework.

    \item We systematically re-execute up to 20 state-of-the-art models using their publicly available implementations under equal and controlled experimental conditions, enabling fair and \textbf{reproducible benchmarking} and alleviating the need for future researchers to independently reproduce these baselines for comparison.
    
    \item We \textbf{release an open-source benchmarking framework} to the scientific community, designed to facilitate the evaluation of new approaches and their direct comparison against the most influential models in the field.
\end{itemize}

%% Structure of the paper
The remainder of the paper is structured as follows. Section \ref{sec:search_methodology} describes the literature search methodology adopted to ensure an unbiased review. Section \ref{sec:related_work} presents the related work, covering both early foundational studies and existing frameworks and benchmarking efforts. Section \ref{sec:analysis} introduces the proposed unifying framework and provides a structured analysis and classification of the reviewed approaches. Section \ref{sec:benchmark} details the benchmarking protocol, including the experimental settings and evaluation criteria. Section \ref{sec:results} reports and discusses the results of the empirical study. Finally, Section \ref{sec:conclusion} concludes the paper and outlines future research directions.

\section{Search methodology} %% CHAPTER 2
\label{sec:search_methodology}

The objective of this review is to provide a comprehensive and structured overview of recent advances in toxicity prediction for drug discovery, with particular emphasis on deep learning methods and graph-based models. Unlike traditional systematic literature reviews, toxicity‐related research spans a highly diverse and heterogeneously described landscape: toxicity endpoints vary widely across studies, the terminology used is non-standardized, and relevant works are often indexed under disparate domains such as pharmacology, cheminformatics, systems biology, ADMET modelling, or risk assessment. In practice, this heterogeneity resulted in two major challenges: (i) broad keyword queries returned an unmanageable number of non-relevant publications, and (ii) narrowing the queries led to the exclusion of core papers that did not explicitly use the expected terminology.

Given these domain-specific constraints, a fully systematic protocol —e.g., PRISMA— was not appropriate nor reproducible without substantial loss of relevant works. Instead, this review follows a structured literature mapping approach informed by principles of the PRISMA-ScR (\textit{Scoping Review}) framework \cite{PrismaScR}. Scoping reviews are particularly suited for research areas with diffuse terminology, evolving methodologies, and heterogeneous study designs —conditions that align with the current toxicity prediction landscape—. Following this rationale, we adopted a transparent and methodical process adapted to the characteristics of the field based on the PRISMA-ScR methodology.

\paragraph{\textbf{Search strategy and sources}}

\begin{figure}[tb!]
  \centering
  \includegraphics[width=\linewidth]{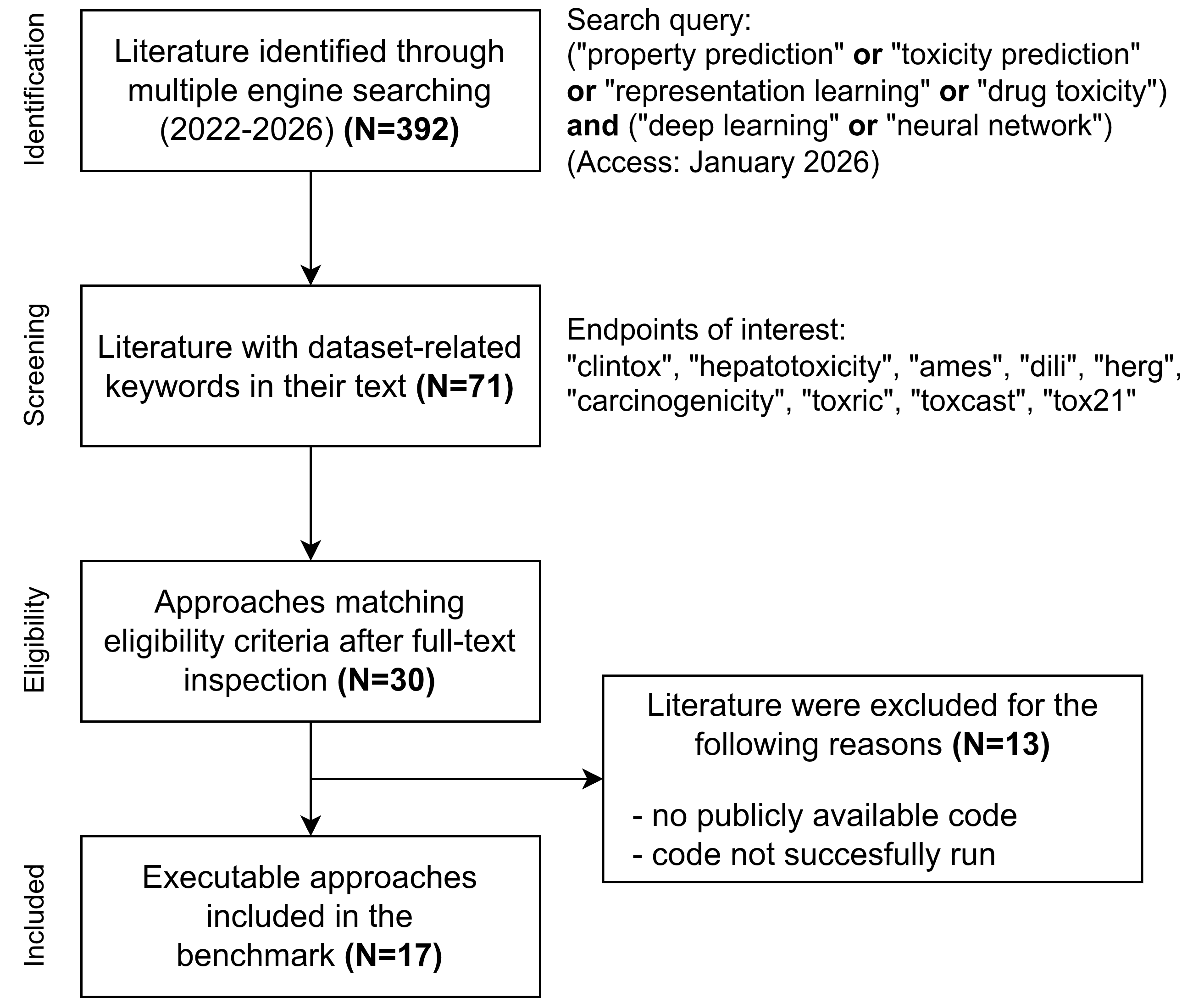}
  \caption{Flow diagram for literature search}
  \label{fig:flow_diagram}
\end{figure}

The literature search covers peer-reviewed work published from  
January 2022 to January 2026 and follows a multi-stage retrieval and filtering process, as illustrated in \autoref{fig:flow_diagram}.

The initial stage employed a keyword-based query across multiple academic databases and publication repositories, including ACM Digital Library, IEEE Xplore, PubMed, ScienceDirect, Scopus, SpringerLink Google Scholar and Linknovate. The search query combined task-related terms with modelling-oriented terminology as can be seen on the figure.
The results of the query came to almost 400 records after deduplication. %as follows:

In the second stage, the retrieved records were screened for the presence of dataset-related keywords in the full text. Specifically, studies were retained if they referenced any of the following widely adopted toxicity benchmark datasets or endpoints: \textit{clintox}, \textit{ames}, \textit{hepatotoxicity}, \textit{dili}, \textit{herg}, \textit{carcinogenicity}, \textit{toxic}, \textit{toxcast}, or \textit{tox21}. This filtering step yielded 71 papers.

\paragraph{\textbf{Eligibility criteria}} The third stage consisted of a full-text eligibility assessment. Studies were included only if they met the following criteria: (i) the proposed approach fell within the scope of graph-based deep learning methods for toxicity prediction in drug discovery —excluding works leveraging different methods or focusing exclusively on environmental toxicity, ecotoxicology, regulatory toxicology or wet-lab assays without computational modelling—, and (ii) the study included an original model proposal for in-silico prediction, leaving apart literature reviews or comparative works. This cutted by half the articles that made it to the review.

%Works were discarded if they met at least one of the following exclusion criteria: 
%\begin{itemize}
    %\item Studies not fitting any of the considered categories, focusing exclusively on environmental toxicity, ecotoxicology, regulatory toxicology or wet-lab assays without computational modelling.
    %\item Studies that did not include an original model proposal, such as literature reviews or comparative works.
%\end{itemize}

Lastly, a final stage was conducted to select those approaches with fully executable codebases and reproducible results to take part in the benchmark. This is detailed in the corresponding section later in this work.

\section{Related work} %% CHAPTER 3
\label{sec:related_work}

Early applications of deep learning to computational toxicity prediction can be traced back to DeepTox \cite{DeepTox}, which demonstrated that multi‑task deep neural networks trained on high‑dimensional molecular descriptors could substantially outperform traditional machine‑learning approaches in the Tox21 Data Challenge \cite{Tox21}. This work established that hierarchical feature learning and end‑to‑end optimization were particularly well suited to capturing complex structure–activity relationships, and it effectively positioned deep learning as a new paradigm for in silico toxicity assessment.

In the years that followed, a first wave of deep models for molecular property prediction began to move away from hand‑crafted descriptors toward learned representations, including early convolutional and recurrent architectures operating on SMILES strings and graph‑based networks that treated molecules explicitly as labeled graphs. These models provided important baselines for toxicity prediction tasks and showed that learned representations could transfer across endpoints, inspiring a series of architectures that progressively tightened the coupling between chemical structure and predictive performance.

A major milestone in graph‑based molecular modeling was AttentiveFP \cite{AttentiveFP} introduced an attention‑based message‑passing framework in which both atom‑wise and molecule‑level attention mechanisms were used to weight the contribution of different substructures, yielding interpretable and highly expressive molecular fingerprints. This architecture achieved competitive performance on multiple toxicity datasets and helped establish attention‑enhanced graph neural networks as reference baselines for subsequent work, particularly in settings where model interpretability and substructure attribution are critical.

Around the same period, GROVER \cite{Grover} introduced graph transformers combining message‑passing with a self‑supervised pretraining on large unlabeled corpora of molecular graphs to learn rich, task‑agnostic node and graph embeddings. By jointly leveraging node‑level and edge‑level contextual information, GROVER set strong baselines on a variety of molecular property benchmarks, including toxicity‑related endpoints, and popularized pretraining and fine‑tuning pipelines that later became standard in the field. An example of this is SSL-GCN \cite{SSL-GCN}, explicitly tailored for toxicity prediction, leveraging self-supervised objectives to better capture structure–toxicity relationships in molecular graphs.

Building on these foundations, subsequent work increasingly focused on enriching molecular graphs with domain knowledge \cite{KPGT, KANO, PharmaHGT, DeepPK} and 3D structural information \cite{GraphMVP, GEM, DumplingGNN, ET-Tox, SynthMol, FATE-Tox}, while also exploiting shared representations across multiple tasks \cite{MMGIN, MPCD} and more modern networks \cite{GIN, LiGhT} and training strategies \cite{MolCLR, CD-MVGNN, 3MTox}. These approaches aimed to better capture complex structure–property relationships and further improve upon established toxicity prediction performance.

As all these architectures proliferated, a growing need emerged for standardized benchmarks and curated datasets to enable fair and reproducible comparison across toxicity prediction methods. Many of the cited models compare themselves using the established benchmark for molecular property prediction MoleculeNet \cite{MoleculeNet}, released as part of the DeepChem library \cite{DeepChem}. MoleculeNet represented a major step forward in standardizing dataset curation and evaluation protocols for a broad range of molecular machine learning tasks. It includes a variety of physicochemical, biochemical, and bioactivity datasets, establishing unified metrics and splitting strategies that have become widely adopted in the field. However, despite its breadth, MoleculeNet was not specifically designed to address toxicity prediction and therefore provides only a limited representation of toxicological endpoints.

Beyond standardized benchmarks like MoleculeNet, several recent reviews \cite{rev_Amorim, rev_Cavassotto, rev_VanTran} have sought to synthesize progress in computational toxicity prediction by compiling results across a diverse range of classical and modern methods, typically covering a broad spectrum of algorithms such as random forests, support vector machines, gradient boosting and generic neural network architectures, along with relevant proposals. These works usually compare models indirectly, by summarizing and juxtaposing the performance metrics reported in the original publications, often across heterogeneous datasets and evaluation protocols. In contrast, the present work goes beyond a narrative review by introducing a dedicated benchmarking framework that focuses on graph-based deep learning for molecular toxicity prediction and evaluates all considered models under a unified experimental setup, ensuring a fair and controlled comparison across architectures.

More recently, dedicated resources such as TOXRIC \cite{Toxric} have addressed the need for clean and standard data sources of these machine learning methods. TOXRIC is a recently introduced resource that combines a comprehensive toxicology database with machine-learning–ready datasets and baseline benchmarks, providing an extensive suite of compounds from different toxicity categories and endpoints that greatly facilitates data access and standardization. Following its release, many studies in computational toxicity prediction  \cite{Nyan_reuse, MMGIN, muToxAl} have adopted its curated data resources for both training and evaluation. Our contribution likewise builds on the TOXRIC versions of several widely used toxicity datasets, which we select as the core of our benchmark. Apart from data curation, TOXRIC offers baseline benchmarks constructed by systematically combining multiple feature types with a small set of typical machine learning algorithms, together with visualization tools to inspect molecular representations and benchmark results for each endpoint.

Complementary to resource-oriented efforts such as TOXRIC, our work focuses on the systematic evaluation of state-of-the-art graph-based models for molecular toxicity prediction under a unified and reproducible framework. Rather than exploring combinations of generic feature representations and baseline algorithms, we design a specialized benchmark that standardizes datasets, endpoints, data splitting strategies, and evaluation protocols. This enables rigorous, head-to-head comparisons across architectures while reducing variability introduced by inconsistent experimental setups.

Specifically, our benchmark builds upon curated TOXRIC datasets and extends prior initiatives such as MoleculeNet by incorporating a broader and more representative set of toxicological endpoints, alongside additional splitting strategies that better reflect real-world generalization scenarios. We further provide open-source implementations, detailed experimental protocols, and comprehensive results for all evaluated models, ensuring transparency and facilitating straightforward reuse.

Overall, this work establishes a robust and extensible benchmarking framework tailored to graph-based toxicity prediction, addressing current limitations in comparability and reproducibility. By consolidating modern architectures within a controlled experimental setting, it provides a reliable foundation for future methodological advances and fair performance assessment in computational toxicology.

%%% ===============================================
%%%  START OF SECTION 
%%% ===============================================

\section{Analysis and classification} %% CHAPTER 4
\label{sec:analysis}

\begin{figure*}[tb!]
  \centering
  \includegraphics[width=\linewidth]{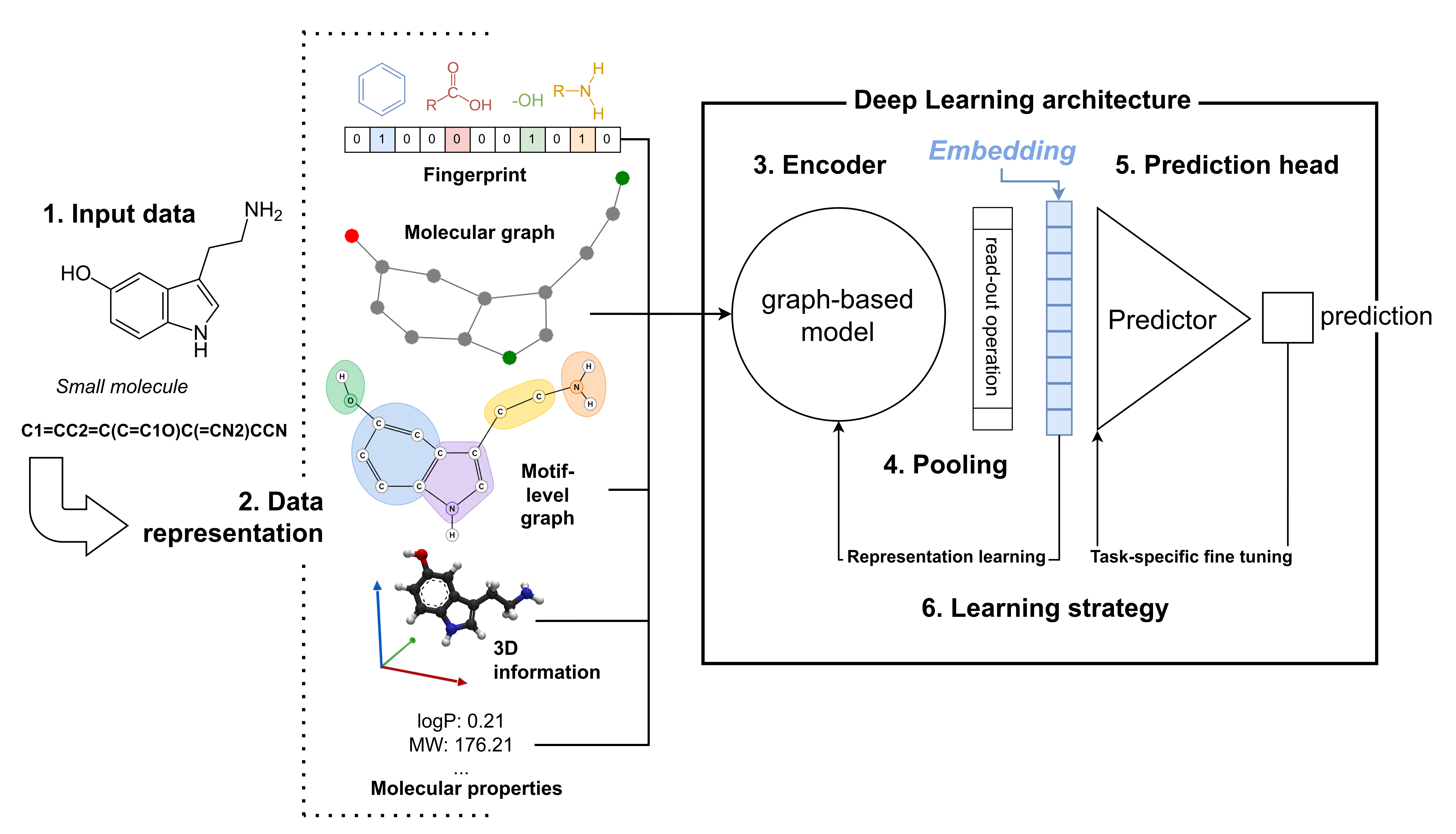}
  \caption{Components of the property prediction general framework. \textbf{1.} Small molecule provided in SMILES format \textbf{2.} Different data  representations that can be inputed to the deep learning model as complementary information or separate views in multimodal settings \textbf{3.} Graph-based neural model that takes at least one graph as input and generates a dense representation for the provided features \textbf{4.} Pooling mechanism used to summarize the node-level representations into a single embedding representing the molecule \textbf{5.} Neural network acting as prediction head for generating the target value from a deep representation of the molecule containing presumably all relevant information for the task \textbf{6.} Method applied to optimize the deep learning model for the task at hand. Different learning strategies can be leveraged for this purpose, such as dividing training into representation learning and fine tuning stages.}
  \label{fig:framework}
\end{figure*}

In \autoref{fig:framework}, we provide an overview of the general framework underlying the works reviewed in this manuscript. The problem is formulated starting from a small molecule, typically represented in SMILES format, for which an unknown property is of interest, in this case toxicity.

This molecule can be represented in multiple ways. The most popular are \textbf{fingerprints}, a binary vector descriptor that captures information about the molecule’s structure and functional groups, and \textbf{molecular graphs}, where each node corresponds to an atom and each edge to a bond. It is also common for approaches to incorporate additional knowledge in the form of a higher-level graph encoding functional groups and strucutal motifs, 3D information, or other known molecular properties.

A subset of this information is fed into a deep learning model, typically graph-based to align with the molecule’s nature. In multimodal models that integrate different types of information, each view is usually processed in a separate branch with its own encoder, and the branches are then combined via concatenation or another fusion mechanism. A \textbf{pooling operation} is commonly required to move from a fine-grained representation —at the atomic or functional level— to a single \textbf{molecular embedding}. Once we have this dense representation, a simple predictor network is usually employed to obtain the target value.

During training, these models are optimized to correctly predict toxicity values for a known set of molecules —the training set— and validated on a set of molecules not seen during training —the test set—. Many approaches separate the training process into two stages. First, a pretraining stage on a very large set of molecules to learn to represent this data effectively, typically by optimizing a generic or self-supervised task to obtain a representative embedding —in a task called \textbf{representation learning}—. Then, this model is \textbf{fine-tuned} on a task-specific toxicity dataset, smaller in size, with the goal of achieving accurate predictions by leveraging prior domain knowledge.

In this section, we detail the different components employed by the reviewed architectures. A structured summary of how these components are implemented across approaches can be found in \autoref{tab:approaches}. Each subsection further examines the individual components, outlining the design choices adopted in the literature.

%% Sumary table
\begin{table*}[tb!] %
    \centering
    \caption{Classification of reviewed approaches}
    \label{tab:approaches}
    \small
    \setlength{\tabcolsep}{3pt}
    \renewcommand{\arraystretch}{1.0}
    
    \begin{tabular}{c l l l l l l r}

    \toprule
    \textbf{Ref.} & \textbf{Approach} & \textbf{Input} & \textbf{Data Representation} & \textbf{Encoder} & \textbf{Pooling} & \textbf{Learning Strategy} & \textbf{Primary Task} \\
    \midrule

    \cite{AttentiveFP} & AttentiveFP & 2D & AG & GAT & GRU & Classical & MPP \\

    \cite{Grover} & Grover & 2D & AG + K & GT & Attention & SSL & MPP \\

    \cite{SSL-GCN} & SSL-GCN & 2D & AG & GCN & Max-pool. & SSL & TP \\

    \cite{MolCLR} & MolCLR & 2D & AG & GCN + GIN & GAP & SSL, Contrastive & MPP \\

    \cite{KPGT} & KPGT & 2D & \textbf{LG} + FP & LiGhT & GAP & SSL & MPP \\

    \cite{GEM} & GEM & \textbf{3D} & AG + \textbf{LG} & GIN (GeoGNN) & GAP & SSL & MPP \\

    \cite{CD-MVGNN} & CD-MVGNN & 2D & AG + \textbf{LG} & NodeGNN, EdgeGNN & Attention & Disagreement loss & MPP \\

    \cite{GraphMVP} & GraphMVP & \textbf{3D} & AG & VGAE & - & SSL & MPP \\

    \cite{RG-MPNN} & RG-MPNN & 2D & AG + \textit{MG} & GAT & Attention & Classical & MPP \\

    \cite{Nyan_reuse} & NYAN & 2D & AG + FP & VGAE + ExtraTree & - & Pretrain & MPP \\

    \cite{PharmaHGT} & PharmaHGT & 2D & AG + \textit{MG} & HGT & GRU + Attention & Classical & MPP \\

    \cite{KANO} & KANO & 2D & AG + KG & CMPNN & Max-pool., GRU & Contrastive, Prompt & MPP \\

    \cite{GeoDILI} & GeoDILI & \textbf{3D} & AG + \textbf{LG} & GIN (GeoGNN) & GAP & Pretrain & DS \\

    \cite{ET-Tox} & ET-Tox & \textbf{3D} & AG & Equivariant GT & Attention & Classical & TP \\

    \cite{FSGNNTR} & FS-GNNTR & 2D & AG & GIN + Transformer & GAP & Pretrain & MPP \\

    \cite{ToxMPNN} & ToxMPNN & 2D & AG & Gated GNNs & Attention, Max-pool. & Classical & TP \\

    \cite{MMGIN} & MMGIN & 2D & AG + FP & GIN & GMP & Multi-task & TP \\

    \cite{3MTox} & 3MTox & 2D & \textbf{LG}/\textit{MG} + FP & Transformer & [CLS] & SSL, Contrastive & TP \\

    \cite{MPCD} & MPCD & 2D & AG & GT & Attention & Multi-task & MPP \\

    \cite{AttenhERG} & AttenhERG & 2D & AG & Attention & Attention & Classical & DS \\

    \cite{DeepPK} & DeepPK & 2D & AG + FP + K & D-MPNN & - & Classical & TP \\

    \cite{MolPROP} & MolProp & 2D & AG + SMILES & GAT + BERT & Attention & Pretrain & MPP \\

    \cite{hERGAT} & hERGAT & 2D & AG + FP + feat. & GAT + GRU & Attention + GRU & Classical & DS \\

    \cite{DILI_GATNN} & DILI\_GATNN & 2D & AG + FP & GAT + DNN & GAP & Pretrain & DS \\

    \cite{DumplingGNN} & DumplingGNN & \textbf{3D} & AG & GAT + GraphSAGE & GAP & Classical & MPP\\

    \cite{SynthMol} & SynthMol & \textbf{3D} & AG + FP & GAT + UniMol & GRU & Pretrain & MPP \\

    \cite{FATE-Tox} & FATE-Tox & \textbf{3D} & AG + \textit{MG} + FP & SE(3)-Transformer & Sum-pooling & Classical, Multi-task & TP \\

    \cite{SamarMonem} & Samar Monem & 2D & AG & Attention & Concat, Attention & Classical & TP \\

    \cite{ToxKG} & ToxKG & 2D & KG + FP & Heterogeneous GraphGPS & - & Classical & TP \\

    \cite{Mamba} & AMPred-LWN & 2D & AG + \textit{MG} + FP & GAT, GIN, Mamba & Mamba & Pretrain & DS \\

    \bottomrule
    \end{tabular}
\captionsetup{font=footnotesize}
\caption*{
Approaches are presented in chronological order. \textbf{\textit{Abbr.:}}  AG = Atom Graph, \textbf{LG} = Line Graph, \textit{MG} = Motif Graph, K = Knowledge, KG = Knowledge Graph, FP = Fingerprint. 
MPP = Molecular Property Prediction, TP = Toxicity Prediction, DS = Domain-Specific tasks such as hepatotoxicity or cardiotoxicity.}
\end{table*}

\subsection{Primary task}

First, it is necessary to frame the reviewed architectures according to the primary task or application domain they target. This perspective provides a high-level categorization that helps contextualize the different design choices discussed throughout this section, distinguishing between general-purpose molecular representation learning, toxicity-focused models, and domain-specific approaches.

\subsubsection{Molecular property prediction}
This line of work focuses on molecular encoders that learn general-purpose representations of small molecules, independently of any single downstream endpoint. These models are usually evaluated on broad benchmarks that include diverse physicochemical and bioactivity tasks, sometimes with a subset of toxicity-related datasets, like MoleculeNet. Their main goal is to produce transferable embeddings that can be reused across multiple prediction problems with minimal task-specific adaptation.

\subsubsection{Toxicity prediction}
Toxicity prediction can be viewed as a particular case of molecular property prediction, but it is often treated as a distinct task due to its central role in risk assessment and drug safety. Within this setting, models are explicitly optimized to distinguish between toxic and non-toxic compounds or to estimate toxicity-related endpoints, frequently in the context of ADMET profiling, where safety and efficacy must be balanced simultaneously. Compared with general molecular property models, these approaches typically focus on toxicity-relevant endpoints and design choices that favour interpretability and reliability in safety-critical scenarios.

\subsubsection{Domain-specific toxicity}
Toxicity covers a broad spectrum of effects, from organ-specific and systemic toxicity to ecological and environmental endpoints. As a result, there is an increasing interest in domain-specific models tailored to particular forms of toxicity, such as hepatotoxicity, cardiotoxicity, or acute and ecotoxic effects. These models are usually trained and evaluated on specialized datasets and incorporate inductive biases or features that reflect the underlying biological mechanisms of the target endpoint, trading some degree of generality for improved performance and relevance within their specific domain.

Although these models are typically evaluated on a single type of toxicity using specialized datasets, some architectures exhibit the potential to generalize beyond their original domain. For this reason, we also consider them within our benchmark, assessing their performance against other state-of-the-art approaches across a broader range of toxicity prediction tasks.

\vspace{1em}
This task-oriented categorization establishes the context in which architectural design choices should be interpreted, and guides the component-wise analysis that follows.

\subsection{Input data}

Molecular features form the essential interface between chemical structures and computational toxicity prediction models, encoding molecular information in formats that preserve physicochemical and topological properties while enabling efficient machine learning. These features are systematically classified by dimensionality, beginning with \textbf{zero-dimensional (0D)} descriptors that capture only constitutional information —such as molecular weight, atom counts, and basic elemental composition— offering simplicity but limited structural insight. \textbf{One-dimensional (1D}) descriptors extend this by incorporating linear substructural elements, including functional group frequencies, ring counts, and chain lengths, often represented through string-based notations like SMILES, that sequences atoms and bonds but remain sensitive to ordering conventions \cite{DDREV}.

\textbf{Two-dimensional (2D)} descriptors provide substantially richer information by modeling molecules as undirected graphs, thereby encoding adjacency, connectivity, and topological features through graph invariants, connectivity indices, and substructure-based fingerprints. \textbf{Three-dimensional (3D)} descriptors further enhance expressiveness by incorporating spatial coordinates for each atom, enabling the quantification of steric effects, pharmacophoric arrangements, molecular volume, and surface properties that are crucial for predicting receptor interactions, metabolic liabilities, and organ-specific toxicities, albeit at the cost of conformational sampling \cite{DDREV}.

This progression from 0D/1D simplicity toward 2D topological and 3D geometric fidelity directly supports the requirements of graph neural networks, which natively process molecular graphs and conformers to learn adaptive, hierarchical embeddings that surpass the limitations of fixed-length descriptors. The predominance of 2D graph-based representations in modern toxicity prediction frameworks, sometimes enriched with 3D information, is coherent with the choice of graph deep learning architectures that are the focus of this review.

\subsection{Data representation}

Molecules can be represented in multiple ways, each providing a different view of the same underlying structure, depending on which features are deemed most relevant for the task. Some of these representations are complementary, while others capture fundamentally different perspectives.

\subsubsection{Molecular fingerprints}

Molecular fingerprints are a specific class of molecular descriptors that encode chemical structures into fixed‑length vectors that can be directly processed by machine‑learning models. Depending on the implementation, these vectors can be purely binary —indicating presence/absence of features— or contain integer counts of fragments or motifs, providing more nuanced information on feature multiplicity. Within the broader context of domain knowledge and molecular representation, fingerprints thus act as a bridge between structural chemistry and statistical learning by transforming 2D or 3D structural information into machine‑readable features that have been widely used in QSAR and read‑across approaches for chemical safety assessment.

    \textbf{Classical key‑based fingerprints} such as MACCS encode the presence or absence of a predefined set of substructures —e.g., specific ring systems, heteroatoms, functional groups— into fixed‑length binary vectors \cite{MACCS}. These fingerprints are computed by matching each key pattern against the molecular graph and setting the corresponding bits, which makes them highly interpretable and efficient for similarity searching, diversity analysis and as a baseline representation in toxicity QSAR models. For instance, MACCS keys combined with machine‑learning algorithms have been used to predict acute oral toxicity with good performance \cite{GoodMACCS}. \textbf{Dictionary‑ or path‑based fingerprints} like the PubChem 2D fingerprint \cite{PubchemFP} or the RDKit substructure/path fingerprints \cite{RdkitFP} follow a similar principle but rely on larger and often more exhaustive sets of paths or fragments, yielding longer bitstrings with broader coverage of structural motifs that are particularly suitable for virtual screening, clustering and similarity‑based read‑across in toxicology datasets.

\textbf{Circular fingerprints} such as ECFP \cite{ECFPs} and Morgan fingerprint \cite{MorganFP} encode local atomic environments by iteratively hashing atom‑centered neighborhoods up to a specified radius. In this scheme, each atom’s environs —defined by atom type, neighboring atoms, bond orders and sometimes additional features— are converted into identifiers that are finally folded into a binary or count vector. Since they highlight local substituent patterns and their connectivity, circular fingerprints are particularly effective for capturing structure–toxicity relationships, supporting high‑performing classification and regression models, scaffold hopping and detection of subtle toxicophoric patterns that may not be captured by simple substructure keys  \cite{BenchmarkFPs}. \textbf{Avalon fingerprints} implement a flexible, hash‑based encoding of paths and features that can be configured in size and detail, generating binary or count vectors from combinations of atom and bond features. This tunability makes Avalon attractive for benchmarking models across multiple representation sizes or when optimizing trade‑offs between dimensionality and predictive performance. Furthermore, past research has shown that Avalon molecular fingerprints excel in multi-endpoint acute toxicity tasks \cite{Nyan_reuse}.

\textbf{Pharmacophore‑oriented fingerprints}, including ErG‑type encodings \cite{ErG_FP, PharmacophoreErG}, represent molecules in terms of pharmacophoric features such as hydrogen‑bond donors and acceptors, aromatic centers, charged groups and hydrophobic regions, along with their topological or 3D distances \cite{PharmacophoreFP}. These fingerprints are typically derived from conformer ensembles or pseudo‑3D representations and are particularly well suited for modeling endpoints that depend on specific spatial arrangements of features —e.g., target‑mediated toxicity or off‑target binding—, as illustrated by ErG‑based models that successfully classify ligands for E3 ligases and other pharmacological targets \cite{PharmacophoreErG_e3, ErG_PKL1}.

In the context of toxicity prediction, the choice of fingerprint determines which aspects of chemical space and mechanistic information are emphasized. Key‑based and dictionary fingerprints are advantageous when interpretability, substructure alerts and fast similarity search are priorities, for example to rationalize structural alerts associated with mutagenicity or hepatotoxicity \cite{ExplainableFP}. Circular fingerprints often deliver superior predictive performance in machine‑learning benchmarks \cite{BenchmarkFPs}, making them a common default for large‑scale toxicity classification and regression tasks. Pharmacophore and 3D‑aware fingerprints become particularly relevant when toxicity is mediated by well‑defined ligand–target interactions where spatial feature arrangements are critical. Approaches such as MMGIN \cite{MMGIN} and SynthMol \cite{SynthMol} integrate multiple complementary fingerprint representations —including MACCS, pharmacophore ErG, and PubChem fingerprints— in an attempt to capture diverse molecular features and enhance the informational richness available to the model.

Even so, all fingerprints are expert-designed automatically-generated representations that capture incomplete or partial information about the molecule that might not be relevant or optimal for the problem at hand. Despite their popularity and utility, specially by facilitating the incoporation of domain-specific expertise unknown to computer scientists, these techniques often fall short to archive the desired predictive perfomance for challenging tasks. For this reason, current efforts are directed toward generating richer context-aware representations through dynamic, self-learned embeddings that move beyond static molecular encodings \cite{AttentiveFP, LearnedFPs}.

\subsubsection{Molecular graphs}

Molecules are commonly represented as \textbf{atom-centered graphs}, where nodes correspond to atoms and edges to chemical bonds \cite{SSL-GCN, GraphMVP, MolCLR, FS-GNNTR, SamarMonem}. In this representation, both nodes and edges are associated with feature vectors encoding physicochemical properties, such as atom type, hybridization state, formal charge, or bond type and aromaticity. These features provide the basis for message passing in graph-based models.

\begin{figure}[tb!]
  \centering
  \includegraphics[width=\linewidth]{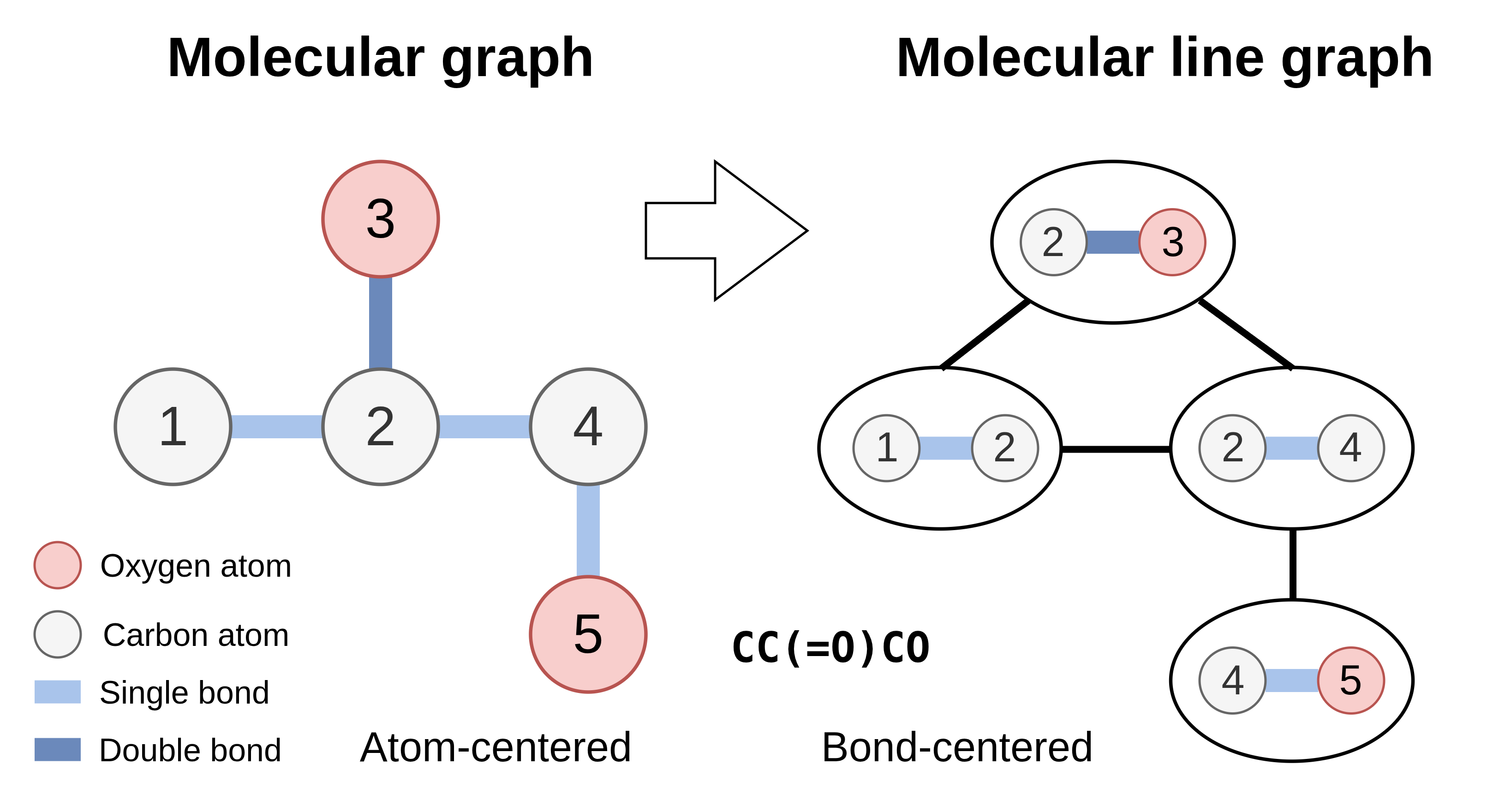}
  \caption{An illustrative example of the transformation of a molecular graph to a molecular line graph \cite{KPGT}.}
  \label{fig:line_graph}
\end{figure}

An alternative formulation focuses on \textbf{bond-centered representations} through the use of line graphs \cite{LiGhT}, where nodes represent bonds and edges encode adjacency between bonds \cite{KPGT, GEM, CD-MVGNN}. As illustrated in \autoref{fig:line_graph}, this transformation shifts the modeling perspective from atoms to interactions between bonds, enabling the model to more explicitly capture patterns related to bond environments, connectivity, and local structural arrangements. This is particularly relevant for toxicity prediction, where many mechanisms are driven by specific substructures or reactive configurations —such as conjugated systems or electrophilic groups— that are more naturally characterized at the bond level than at the level of isolated atoms.

%%%  ATOM-BASED V.S. BOND-BASED GRAPH

The new graph will have a node for each pair of bonded atoms and edges between nodes sharing atoms. This type of representation is leveraged by KPGT \cite{KPGT} and, implicitly, by other approaches such as CD-MVGNN \cite{CD-MVGNN}, GEM \cite{GEM}, GeoDILI \cite{GeoDILI} and 3MTox \cite{3MTox}, which incorporate bond-level graphs into their modelling. By operating on bonds, these methods can better model how local structural contexts influence chemical behavior, leading to more informative representations for downstream prediction tasks.

Furthermore, bond-centered modeling facilitates the incorporation of geometric information \cite{LiGhT}. For example, GEM and GeoDILI leverage this representation to include bond angles as edge features, since angles are defined between pairs of adjacent bonds \cite{GEM}. This provides a natural mechanism to encode 3D structural information, allowing the model to account for molecular geometry and spatial constraints that are often critical for accurately predicting toxicity \cite{GeoDILI}.

\subsubsection{Motif-level graphs}

Some approaches incorporate domain knowledge by introducing higher-level graph representations that capture structural motifs, in an attempt to approximate functional groups and toxicophores.  These motif-level graphs aim to abstract recurring chemical patterns that are often associated with specific biological or toxicological properties.

There are multiple strategies to construct such representations. Some methods rely on predefined lists of substructures, as in AMPred-LWN \cite{AMPred-LWN}, RG-MPNN \cite{RG-MPNN} or DeepPK \cite{DeepPK}, to identify known functional groups or toxicophores. Others apply algorithmic decomposition techniques, such as BRICS fragmentation \cite{BRICS} in PharmHGT \cite{PharmaHGT}, to partition molecules into chemically meaningful components. Alternative approaches, such as 3MTox \cite{3MTox}, focus on specific structural elements like ring systems. By operating at this higher level of abstraction, these models can capture semantically meaningful patterns that may be less evident at the atom level.

\subsubsection{3D information}

Beyond topological representations, the three-dimensional (3D) structure of molecules plays a crucial role in determining their properties, including toxicity. This additional information can significantly enhance the model’s ability to learn structure–property relationships.

Molecular geometry arises from the balance of attractive and repulsive forces between atoms, leading to specific spatial conformations. In practice, molecules can adopt multiple conformations —known as conformers— which correspond to different local minima in the energy landscape. These conformers can be generated using computational methods such as force-field optimization or more accurate quantum chemical calculations.

Some approaches explicitly incorporate 3D coordinates or distance-based features into graph models, enabling the capture of geometric relationships such as interatomic distances and angles. This is the case from GEM, GeoDILI, GraphMVP \cite{GraphMVP}, ET-Tox \cite{ET-Tox}, DumplingGNN \cite{DumplingGNN}, FATE-Tox \cite{FATE-Tox} and SynthMol \cite{SynthMol}, which does so by leveraging pre-trained embeddings from the Uni-Mol framework \cite{Unimol}.

\subsubsection{Molecular properties}

In addition to structural representations, models may incorporate global molecular descriptors, which encode known physicochemical properties at the molecule level. These include features such as molecular weight (MW), octanol–water partition coefficient (logP), topological polar surface area (TPSA), or the number of hydrogen bond donors and acceptors. Such descriptors, often referred to as 0D representations, provide complementary information that can be directly related to toxicity, for example through their influence on bioavailability, membrane permeability, or reactivity, all of which are closely connected to toxicological behavior.

One approach that incorporates this type of information is hERGAT \cite{hERGAT}, which has been proposed for hERG blockade prediction, a setting in which cardiotoxicity is strongly influenced by global physicochemical properties and substructural features associated with channel inhibition.

This type of information has also been incorporated in  for hERG blockade prediction, a setting in which cardiotoxicity is strongly linked to molecular properties and substructural determinants of channel inhibition \cite{hERGAT}.

There are also approaches that explore 1D representations, such as SMILES strings, by processing them using sequential or convolutional architectures. However, these methods fall outside the scope of this work, as we focus on approaches that leverage richer 2D and 3D representations incorporating topological and geometric information, which are particularly relevant for molecular toxicity prediction. In this context, MolPROP \cite{MolPROP} is a representative multimodal approach that combines a graph-based molecular representation with a SMILES-based sequence branch encoded using ChemBERTa \cite{ChemBERTA}, thereby integrating complementary structural and SMILES-derived embeddings for property prediction.

%%%%%%%%%%%%%%%%%%%%%%%%%%%%%%%%%%%%%%%%%
%%%%%%% Neural architectures analysis
%%%%%%%%%%%%%%%%%%%%%%%%%%%%%%%%%%%%%%%%%

\subsection{Encoder model}

The choice of encoder depends on the type of molecular representation and the specific view it provides. For fixed-size representations such as molecular fingerprints or handcrafted molecular descriptors, simple \textbf{Feed-Forward Networks (FFNs)} are typically employed, as they can effectively transform these vectorized inputs into dense latent representations \cite{DNNforQSAR}.

In contrast, when molecules are represented as graphs,\textbf{ Graph Neural Networks (GNNs)} have emerged as a powerful framework for learning over such structured data, enabling the extraction of complex relational and structural information.

\subsubsection{Message-Passing Neural Networks}

Message-Passing Neural Networks (MPNN) are the formal framework that unifies virtually all GNN architectures \cite{MPNN, GraphSage}. The idea is simple: each node sends "messages" to its neighbors, collects messages from its neighbors, and uses them to update its own representation. By this simple trick, the framework enforces an inductive bias inside the network and injects the prior knowledge that "connected things are related".

MPNNs can be dissected into three operations. First, the $\mathrm{MESSAGE}$ function (\ref{eq:message}) creates a "message" from neighbour $u$ to target $v$ using their features and edge attributes. Then, the $\mathrm{AGGREGATE}$ function (\ref{eq:aggregate}) combines all incoming messages. This operation must be permutation-invariant, such as the sum, mean, max or some attention-based operation. Last, the $\mathrm{UPDATE}$ function (\ref{eq:update}) merges the aggregated message with node's current features to obtain an updated representation. We can formalize the procedure with the following equations:

\begin{equation}
m_{v \leftarrow u} = \mathrm{MESSAGE}\!\left(
h_u^{(l)},\, h_v^{(l)},\, e_{uv}
\right)
\label{eq:message}
\end{equation}

\begin{equation}
M_v = \mathrm{AGGREGATE}\!\left(
\left\{ m_{v \leftarrow u} \mid u \in \mathcal{N}(v) \right\}
\right)
\label{eq:aggregate}
\end{equation}

\begin{equation}
h_v^{(l+1)} = \mathrm{UPDATE}\!\left(
h_v^{(l)},\, M_v
\right)
\label{eq:update}
\end{equation}

where $h_u^{(l)}$ and $h_v^{(l)}$ are the hidden representation for nodes $u$ and $v$ in the $l$-th layer, $e_{uv}$ corresponds to the features for the $u \rightarrow v$ edge, $\mathcal{N}(v)$ is the local set of neighbours of $v$ and $^{(l)}$ refers to the current layer in the network.

Every GNN architecture follows this pattern and divert only in how they implement these fundamental operations.

\subsubsection{Graph Convolutional Network}

Graph Convolutional Networks (GCN) were the first foundational model for modern GNNs \cite{GCN}. In this implementation, the message is composed of the representation of the node scaled by a normalization factor $1/\sqrt{d_u \cdot d_v}$ where $d_u$ and $d_v$ represent the degree of the nodes, including themselves. This ensures stable learning across graphs with varying degree distributions and prevents high-degree nodes from dominating. The $\mathrm{AGGREGATE}$ function corresponds to the mean operation, and the $\mathrm{UPDATE}$ is carried out through a linear transformation followed by an activation function that introduces nonlinearity, typically ReLU. This is the neural network behind SSL-GCN \cite{SSL-GCN}.

\subsubsection{Graph Isomorphism Network}

The Graph Isomorphism Network (GIN) is designed to be the most expressive GNN possible \cite{GIN}. To this end, node representation is passed as message without any transformation —preserving injectivity— and all messages are summed together in the $\mathrm{AGGREGATE}$ step instead of averaged. This is because the sum preserves multiset cardinality, unlike operations such as mean or max. 

Finally, the aggregated message $M_v$ is added to the central node representation scaled by some factor and a  Multi-Layered Perceptron (MLP) is applied to learn complex relations following $\mathrm{UPDATE} =\mathrm{MLP}\!\left((1 + \epsilon) \cdot h_v +  M_v \right)$. MLP weights are shared among nodes, but not among layers. Parameter $\epsilon$ can be set to a small constant like 0 (in GIN-0) or be a lernable parameter (in GIN-$\epsilon$). The learnable variant allows the model to automatically adjust the relative importance of the node against its neighbors, while maintaining the injectivity property required for maximum discriminative power. Theoretically, this term ensures that GIN can distingish between certain non-isomorphic graph structures that would otherwise yield identical representations, directly contributing to GIN's equivalence to the 1-Weisfeiler-Lehman test in expressive power \cite{GIN}.

The key difference  between GCN and GIN lies in their aggregation schemes and expressive power. While GCN employs a normalized average aggregation, GIN utilizes an additive aggregation. This allows GIN to preserve relevant structural information that may be lost or diluted by the mean and distinguish a broader class of molecules. The great power of GIN motives its use in GEM \cite{GEM} and MMGIN \cite{MMGIN}, whereas MolCLR exploits the complementary characteristics of both architectures \cite{MolCLR}.

\subsubsection{Graph Attention Networks}

Graph attention networks (GAT) extend message passing by learning how much attention each node should pay to its neighbors, instead of treating all neighbors equally. In these models, a learnable attention mechanism assigns a weight to each neighbor, so that messages from more relevant neighbors have a stronger influence on the updated node representation, and the resulting attention coefficients can often be interpreted as a form of importance score \cite{GAT}. The subsequent $\mathrm{UPDATE}$ step can be implemented in different ways, ranging from simple nonlinear activations such as ReLU to more elaborate transformations that combine the attention‑weighted aggregation with the previous node state, a design choice that will be revisited in later architectures.

Attention-based GNNs have become very popular in molecular property and toxicity prediction, and variants of graph attention are employed in models such as AttentiveFP, AttenhERG \cite{AttenhERG}, hERGAT, DILI\_GATNN \cite{DILI_GATNN}, and many more. Althought pure GAT layers are in principle not guaranteed to reach the same level of discriminative power as GINs, they can work better in practice by capturing rich, context-dependent relationships and focusing on the most informative parts of the molecular graph. Many recent architectures therefore combine both paradigms to balance expressive power and flexibility; for example, AMPred-LWN integrates GIN- and GAT-style components within a single model to exploit the advantages of each design \cite{AMPred-LWN}.

\subsubsection{Graph Transformer}

The Graph Transformer (GT) generalizes the popular transformer architecture proposed in \cite{AttentionIsAllYouNeed} to graph-structured data \cite{GraphTransformer}. Analogous to the original formulation, they employ key, query and value learnable projection matrices to map node representations into distinct subspaces that interact within the self-attention mechanism. This family of architectures has been adopted by approaches like GROVER, MPCD \cite{MPCD} and FS-GNNTR \cite{FS-GNNTR}.

This design allows each node in the molecular graph to attend to any other node, rather than being restricted to its local neighborhood, enabling the modeling of long-range interactions between distant parts of the molecule. Structural biases can be incorporated into the attention mechanism to encode relevant graph information, such as shortest-path distances or connectivity patterns, and related ideas have been extended to 3D settings as in FATE-Tox, where spatial relationships between atoms are taken into account. Along similar lines, 3MTox employs a standard transformer encoder augmented with distance-based biases between atom pairs, so that interatomic distances complement the attention scores without altering the core architecture. After the attention operation, a FFN combined with residual connections and normalization is applied to obtain the final node representations.

\subsubsection{Heterogeneous Graph Transformer}

While many molecular graphs are treated as homogeneous —where all nodes and edges share the same semantics— more expressive formulations consider heterogeneous graphs, which include multiple types of nodes and edges representing different entities and relations within a unified structure. This is particularly useful in molecular settings where different components —such as atoms, functional groups, or higher-level motifs— as well as different types of interactions, may carry distinct meanings.

Heterogeneous Graph Transformer (HGT) extend the transformer framework to account for this diversity by adapting message passing to the types of nodes and edges involved \cite{HGT}. Messages exchanged between nodes are modulated by both the type of the source node and the type of the connecting edge, allowing the model to capture more nuanced, relation-specific interactions. In this sense, attention in heterogeneous graphs closely parallels its homogeneous counterpart, but incorporates node-type-dependent weight matrices and edge-type-specific bias terms. Node-type-specific linear transformations and FFNs are further applied to obtain the final representations.

This design enables a more flexible integration of heterogeneous molecular information, leading to richer and more informative representations. PharmHGT   \cite{PharmaHGT} exploits this by using heterogeneous nodes and relations to jointly represent atoms together with higher-level fragments or pharmacophore-like patterns, enriching the molecular graph with multi-granular structural information. In contrast, ToxKG \cite{ToxKG} applies a heterogeneous graph model to knowledge graphs where nodes encode entities such as chemicals, targets, or genes and edges encode curated domain relations, allowing the integration of external toxicological knowledge into the learned representations.

\subsubsection{Line Graph Transformer}

Line Graph Transformer (LiGhT) combines the flexibility of attention-based models with a molecular representation that emphasizes bond connectivity and local context \cite{LiGhT}. This allows the architecture to capture both fine-grained structural cues and more global molecular patterns within a unified framework. In this sense, LiGhT serves as the base architecture featured in KPGT, tailoring the transformer framework to this type of bond-centered input and better exploiting its structural information.

\subsubsection{Variational Graph Autoencoder}

The Variational Graph Autoencoders (VGAE) extends the VAE framework \cite{VAE} to graph-structured data, enabling unsupervised learning of interpretable latent representations for undirected graphs \cite{VGAE}. VGAE learns latent representations $Z$ from a graph with adjacency matrix $A$ and node features $X$, using a GCN-based encoder that parameterizes a multivariate Gaussian posterior and a decoder that reconstructs the adjacency matrix via inner-product similarities. Mean and variance of the latent distribution are obtained from a two-layer GCN that shares weights for the first layer $Z\mu, Z\sigma = \mathrm{GCN}(X, A)$. Then, the decoder reconstructs the adjacency matrix $\hat{A}$ probabilistically from the latent node embeddings $Z$. This architecture has been adopted by NYAN, originally introduced in \cite{Nyan_original} and later revisited in \cite{Nyan_reuse} for toxicity, to obtain a dense molecular representation by jointly encoding the graph structure and fingerprint-based information.

\subsubsection{Multimodality}

Some models simultaneously combine different ways of representing the same molecule, such as graph structures, fingerpints and 3D information within a single architecture. These are commonly referred to as multimodal models, as they leverage multiple complementary views of the data at once. A typical pattern is to process each modality in a dedicated encoder and then fuse the resulting representations through concatenation, attention, or more sophisticated fusion mechanisms.

For example, MMGIN employs a two‑channel architecture that independently processes molecular graphs and fingerprints, learning parallel views of the same compound and later combining them into a unified representation for multitask toxicity prediction. This design allows the model to exploit both topological patterns from the graph and compact, high‑level descriptors from the fingerprint, which can be particularly useful when different toxicity endpoints are governed by different structural cues.

SynthMol adopts a similar multimodal strategy but integrates pre‑trained 3D structural features, graph‑based atom representations, and molecular fingerprints. By leveraging geometric information alongside graph and descriptor inputs, SynthMol is able to capture richer structure–property relationships relevant to drug safety, including effects tied to conformation and spatial arrangement. As in MMGIN, the different modality-specific encoders are ultimately merged via concatenation, preserving the contribution of each source in a single joint representation.

Similarly, AMPred‑LWN is a multimodal multi‑granularity model that fuses atomic‑level graphs, sequences of functional groups, and molecular fingerprints for Ames mutagenicity prediction. It uses enhanced graph neural networks, sequence‑aware Mamba‑based modules, and a dedicated fusion mechanism to adaptively modulate the relative contribution of each branch and combine information across different levels of abstraction, from local atomic neighborhoods to higher‑level functional motifs. Other multimodal toxicity models explore more flexible fusion strategies such as attention-based weighting, exemplified by ToxMPNN \cite{ToxMPNN} and Samar Monem \cite{SamarMonem}.

Altogether, these approaches illustrate how multimodal architectures can exploit diverse sources of molecular information —topological, geometrical, and expert‑handcrafted features— while preserving the strengths of each representation and enabling more robust and interpretable toxicity predictions.

\subsection{Pooling}

Graph pooling in graph neural networks is a critical operation designed to hierarchically reduce the size of a graph by coarsening it into a compact representation while preserving structural and semantic information, tailored to the irregular topology of graphs in contrast to fixed-grid pooling in CNNs.

Crucially, pooling mechanisms must be permutation invariant —meaning they produce identical outputs for isomorphic graphs under any node relabeling— to ensure that representations depend solely on intrinsic graph topology and features, not arbitrary indexing conventions. This property is foundational, as graphs lack canonical node orderings, and permutation equivariance in preceding message-passing layers necessitates invariance in pooling to maintain consistent graph-level predictions across equivalent structures.

Pooling serves dual roles: (i) within $\mathrm{AGGREGATE}$-$\mathrm{UPDATE}$ phases of GNN layers to locally coarsen neighborhoods and mitigate over-smoothing during message passing, and (ii) as a $\mathrm{READOUT}$ function to generate fixed-size embeddings for graph-level prediction. This mechanism enables GNNs to process large-scale graphs efficiently and extract discriminative representations for downstream tasks. Some of the most common operations employed as pooling are:

% Pooling layers are integrated into GNN architectures after message-passing layers to downsample graphs, facilitating hierarchical learning akin to pooling in CNNs. In graph classification tasks, pooling aggregates node representations into a graph-level embedding, often at multiple scales for pyramid-like architectures. It proves particularly valuable in domains like drug discovery—relevant to computational chemistry—where molecular graphs (nodes as atoms, edges as bonds) require coarsening to model toxicity or binding affinity. Applications extend to social networks for community detection, traffic forecasting via road network coarsening, and protein interaction graphs for biological function prediction

\subsubsection{Global Average or Max Pooling}

Global average pooling (GAP) computes a fixed-size vector by averaging node features across the entire graph, offering simplicity, differentiability, and permutation invariance, what makes it popular among multiple approaches. It captures global consensus at the cost of diluting discriminative local signs. Max pooling (MP), conversely, extracts element-wise maxima, emphasizing salient features and exhibiting robustness to noise, often employed in pseudo-max variants for local coarsening. It extracts the most salient motif but ignores broader context that might be relevant. Both are ubiquitous as MPNN aggregators and baseline readouts.
% GAP captures **global consensus** but dilutes discriminative local signals; GMP extracts the **most salient motif** but ignores broader context. Both are ubiquitous as MPNN aggregators and baseline readouts [Gilmer et al., 2017].

\subsubsection{Sequence-based pooling: GRU}

Sequence-based pooling methods, such as those employing Gated Recurrent Units (GRU), Short-Term Memory (LSTM) networks or Mamba \cite{Mamba} \textbf{transform node embeddings into an ordered sequence} —typically via sorting by node degree or a learned permutation— prior to recurrent processing. This approach introduces temporal dynamics to capture hierarchical or sequential patterns within the graph, yielding a permutation-invariant readout through the final hidden state. While effective for graphs exhibiting latent ordering, these methods usually incur higher computational overhead and \textbf{rely on heuristic sorting}, potentially introducing bias depending on the chosen strategy. %in unordered topologies.

We can distinguish between two primary usages of recurrent architectures within graph neural networks. First, they are commonly employed inside the $\mathrm{AGGREGATE}$-$\mathrm{UPDATE}$ step of message-passing layers, where they act as adaptive update functions for node representations. In this setting, the aggregated neighborhood information is first computed through attention or message aggregation mechanisms, for instance as 

\begin{equation}
C_v^{(l)} = \sum_{u \in N(v)} a_{vu}^{(l)} W h_u^{(l)}
\end{equation}

where $a_{vu}^{(l)}$ denotes the attention coefficient between nodes $u$ and $v$, and $W$ is a learnable transformation matrix. Rather than directly replacing the previous node state, the $\mathrm{GRU}$ combines the aggregated context $C_v^{(l)}$ with the prior hidden representation $h_v^{(l)}$ through gated operations:

\begin{equation}
h_v^{(l+1)}=GRU(C_v^{(l)},h_v^{(l)})
\end{equation}

which propagates nonlocal effects while filtering noisy or redundant messages during the aggregation process \cite{AttentiveFP, hERGAT}. By adaptively controlling the amount of newly incorporated information, recurrent updates help stabilize message propagation across multiple layers and preserve relevant contextual dependencies between distant nodes.

Second, recurrent architectures can also be employed as $\mathrm{READOUT}$ operators for graph-level representation learning. In this setting, node embeddings produced after the final message-passing layer are transformed into a sequence —typically following a heuristic or learned ordering strategy— and sequentially processed by the recurrent unit. The final hidden state is then used as a compact graph representation \cite{AttentiveFP, KANO, PharmaHGT, SynthMol}. This enables the model to capture higher-order structural dependencies and hierarchical interactions between nodes, although at the cost of higher computational complexity and sensitivity to the chosen node ordering, which may affect robustness and scalability \cite{HowPowerfulAreGNNs}.

\subsubsection{Attention-based pooling}

Attention-driven pooling employs learnable importance scores to adaptively weight node contributions during coarsening or readout, ensuring permutation invariance through softmax normalization over node scores. Methods such as \textit{DiffPool} and \textit{Set2Set} illustrate two representative designs: \textit{DiffPool} integrates into the $\mathrm{AGGREGATE}$–$\mathrm{UPDATE}$ phases by learning soft cluster assignments that map nodes to supernodes and produce coarsened graphs, whereas \textit{Set2Set} acts as a $\mathrm{READOUT}$ mechanism that iteratively refines a latent state via attention over node features to obtain a global graph representation. Both approaches can highlight salient substructures in molecular graphs, which is particularly relevant in computational toxicology, but their additional parameters and architectural choices often require careful hyperparameter tuning to avoid overfitting.

Several of the reviwed architectures adopt attention-based pooling or readout to focus on toxicity-relevant regions of the molecule. ToxMPNN \cite{ToxMPNN} uses an attention readout over node embeddings to form graph-level representations, enabling the model to emphasize atoms and substructures most associated with toxicity endpoints. AttenhERG, built on the AttentiveFP framework, combines attention along message passing with attention-based readout to improve hERG blockade prediction and to support interpretation by visualizing highly attended atoms and bonds. Models such as MolPROP and hERGAT also rely on attention mechanisms in their graph encoders to compute weighted graph-level summaries. These designs illustrate the variety of strategies —ranging from simple node-level attention pooling to more sophisticated cross-branch attention—  while the multimodal architectures proposed by \cite{SamarMonem} use cross-modal attention to pool and fuse information from different molecular representations in a task-aware mannerthat can be used to derive expressive, toxicity-informed graph representations.

\subsubsection{Virtual node pooling and CLS}

Virtual node pooling augments the graph with a trainable supernode connected to all graph nodes, iteratively updated via message passing or attention to serve as a learnable global aggregator without explicit coarsening. Similarly, CLS token approaches —borrowed from transformer architectures like BERT \cite{BERT}— initialize a dedicated token updated through cross-attention over node embeddings, producing a fixed-size summary that integrates global context in an end-to-end differentiable manner \cite{clspooling}.

These strategies —adopted by models like 3MTox— excel in set-to-vector aggregation for graph-level tasks, maintaining permutation invariance via symmetric aggregation, while preserving expressive power. However, they may underperform on strongly hierarchical structures requiring explicit topology reduction \cite{graphpooling1, graphpooling2}.

\subsection{Prediction head}

Once a graph-level embedding has been obtained, that is, a dense vector representation summarizing the molecule and presumed to concentrate the information most relevant for the task, the most common strategy is to use a FFN network as a prediction head. In practice, this is typically implemented as a MLP that progressively reduces the dimensionality of the embedding through a series of linear transformations and nonlinear activations, ultimately mapping it to either a scalar value or a small output vector corresponding to the target of interest. This design separates representation learning in the GNN backbone from task-specific prediction, and naturally accommodates both single- and multi-task settings.

In relation to the target value that we are trying to predict, problems can be divided into:

\begin{itemize}
\setlength\itemsep{5pt}
\item \textbf{Classification.} These problems aim to predict a discrete label, generally binary, that distinguishes among compounds that have noticeable toxic effects and those considered to be safe. This label is usually calculated using a pre-defined endpoint-specific threshold, derived from expert knowledge, and that is usually unknown for the user of the dataset. 
\item \textbf{Regression.} Here the prediction target comes in the form of a continuous value of a toxic-related metric, such as LD50 (the dose necessary to kill half of the studied population), LDlo (the smallest amount of a substance that has been reported to cause death) or IC50 (the concentration required to inhibit a biological process by half). 
\end{itemize}

For the purpose of this work we have focused in the first kind of problems, as it is the most common approach in the field and most of the reference datasets come in this fashion. Regression is not as frequent since the obtention of such values for humans would require unpracticable experimentation and are usually obtained for other small mammals in a laboratory setting.

However, although the benchmark has been designed for classification tasks with toxicity prediction in mind, the structure and methods employed could be easily extended for regression problems or transferred to other domains, particularly those pursuing some kind of property prediction starting from a small molecule.

\subsection{Learning strategy}

Last but not least, it is also important to characterize how different models are trained and optimized for the given task. This subsection complements the previous components of the framework by emphasizing not only what is being modeled, but also how the corresponding representations are learned.

\subsubsection{Supervised Learning}
Supervised learning remains the cornerstone paradigm for molecular property and toxicity prediction, training graph neural networks on labeled datasets pairing molecular graphs with experimentally measured endpoints such as binding affinities, solubility, or LD50 values. Regression or classification objectives minimize prediction errors via mean squared error or cross-entropy losses, requiring abundant high-quality annotations that are often scarce for rare toxicities.

\subsubsection{Unsupervised or Self-Supervised Learning}
Unsupervised strategies such as VGAE learn latent molecular representations from unlabeled structures by reconstructing graphs, SMILES strings or fingerprints, enabling downstream fine-tuning for property prediction with minimal labels. NYAN follows this paradigm by encoding the molecular graph into a latent space from which it reconstructs the corresponding fingerprint, effectively injecting the information contained in the fingerprint into the learned graph-based representation \cite{Nyan_original}.

On the other hand, self-supervised learning (SSL) defines pretext tasks that do not require manual annotations, such as predicting masked atoms or bonds, distinguishing between real and corrupted subgraphs, or recovering context information from augmented views of the same molecule, thereby encouraging GNN encoders to capture structural invariances that are useful for downstream toxicity prediction. A concrete implementations of this idea is foud in GROVER, which employ node- and graph-level pretext tasks lik motif detection to learn chemically meaningful embeddings that can later be fine-tuned on endpoint-specific toxicity datasets.

\subsubsection{Knowledge-Driven Learning}
Knowledge-driven paradigms infuse domain expertise into molecular learning by incorporating physics-informed losses, quantum-chemical descriptors, or pre-established chemical and pharmacophoric knowledge, particularly in settings with sparse experimental data. Hybrid strategies can enforce chemically motivated constraints or exploit equivariant GNNs that preserve molecular symmetries, guiding optimization toward physically and chemically plausible predictions in extrapolation scenarios, such as the design or assessment of novel toxicants.

Representative examples include KANO \cite{KANO}, which leverages an element- and functional-group-oriented knowledge graph to guide contrastive pretraining and employs functional prompts: molecular embeddings are learned under the guidance of knowledge-derived prompts associated with specific biological or toxicological functions, so that the representation space is shaped along chemically meaningful directions relevant for property and toxicity prediction. KPGT, in turn, introduces an additional knowledge node (K node) for each molecular graph, whose features encode global molecular information such as descriptors and fingerprints. This K node is connected to all other nodes and participates in transformer-based message passing, enabling structural nodes to attend to and exchange information with it, thereby integrating external chemical knowledge into the learned molecular representations.

\subsubsection{Contrastive Learning}

Contrastive learning aims to learn representations by bringing similar (positive) examples closer together in embedding space while pushing dissimilar (negative) examples apart, without requiring explicit labels. In the molecular setting, this is typically achieved by generating different augmented views of the same molecule —e.g., via subgraph perturbations, atom masking, or feature noise— and training the encoder so that embeddings of these views are consistent, while remaining distinguishable from embeddings of other molecules.

MolCLR \cite{MolCLR} instantiates this idea by applying chemically motivated augmentations to molecular graphs —such as atom dropping, bond perturbation, or subgraph masking— and then maximizing the agreement between embeddings of augmented views of the same molecule. This procedure encourages the GNN encoder to capture substructural motifs and global patterns that are stable under reasonable chemical perturbations, which can later be exploited for downstream toxicity prediction in low-label regimes. Along similar lines, KANO combines contrastive learning with a functional prompt mechanism: the model is trained to align molecular embeddings with task- or endpoint-specific prompts, so that the contrastive objective not only enforces consistency across augmented views of a molecule, but also shapes the representation space around functionally  meaningful directions associated with different biological activities or toxicity endpoints.

\subsubsection{Multi-Task Learning}
Multi-Task Learning (MTL) jointly optimizes multiple related endpoints from shared molecular representations, leveraging correlations between tasks to improve generalization and reduce overfitting through a common GNN encoder and task-specific prediction heads. This paradigm is particularly suitable for ADMET profiling and toxicity prediction, where co-training on diverse endpoints can amplify the signal available from sparse labels and produce embeddings that transfer well across related assays. MMGIN, MPCD, and FATE-Tox exemplify this strategy by modeling multiple toxicity- or ADMET-related properties simultaneously from a shared molecular backbone, often achieving better performance than their single-task counterparts while providing a more holistic view of compound safety. %and yielding more robust embeddings that can be transferred across different domains.

%%% ===============================================
%%%  END OF SECTION
%%% ===============================================

\section{Benchmark} %% CHAPTER 5
\label{sec:benchmark}

A major challenge in graph-based molecular toxicity prediction lies not only in the development of increasingly sophisticated architectures, but also in the lack of rigorous and standardized evaluation practices. Existing studies are commonly assessed under heterogeneous experimental conditions, often differing in dataset preprocessing, endpoint selection, train--test splitting strategies, evaluation metrics, number of runs, and statistical validation procedures. Such inconsistencies substantially hinder direct comparison across methods and make it difficult to determine whether reported improvements arise from genuine architectural advances or from differences in experimental design. As a consequence, reproducibility and comparability remain important unresolved issues within the field.

To address these limitations, the principal contribution of this work is the introduction of a unified and reproducible benchmarking framework specifically designed for graph-based deep learning approaches to molecular toxicity prediction. Rather than relying on results reported independently in the literature, we systematically re-evaluate a broad collection of influential state-of-the-art architectures under equal and controlled experimental conditions. This enables fair head-to-head comparisons across models while minimizing confounding factors introduced by inconsistent evaluation protocols.

In contrast to previous benchmarks, our framework builds upon toxicity-specific curated datasets from TOXRIC, chosen to provide a balanced and representative evaluation landscape across diverse toxicological prediction scenarios. Additionally, we assess all models under a unified protocol incorporating several challenging partitioning strategies, together with cross-validation and statistical ranking analyses to ensure robust and meaningful comparisons. Table~\ref{tab:benchmark} summarizes the evaluation methodologies originally employed by the different state-of-the-art approaches and contrasts them with the standardized experimental framework proposed in this work. The full source code, including environment specifications for each model, is publicly available at \url{https://gitlab.citius.gal/noel.suarez/benchtox}.

%% Sumary table
\begin{table}[tb!]
    \centering
    \caption{Original evaluation of benchmarked approaches}
    \label{tab:benchmark}
    \small
    \setlength{\tabcolsep}{3pt}
    \renewcommand{\arraystretch}{1.0}

    \begin{tabular}{l l l l l}

    \toprule
    \textbf{Approach} & \textbf{Datasets} & \textbf{Splits} & \textbf{Val.} & \textbf{Stat. tests} \\
    \midrule

    AttentiveFP & MoleculeNet & \makecell[l]{random,\\ scaffold} & 3-runs & no \\

    Grover & MoleculeNet & scaffold & 3-runs & no \\

    MolCLR & MoleculeNet & scaffold & 3-runs & no \\

    KPGT & MoleculeNet & scaffold & 3-runs & no \\

    GEM & MoleculeNet & scaffold & 4-runs & no \\

    CD-MVGNN & MoleculeNet & scaffold & 10-runs & no \\

    NYAN & TOXRIC & random & 5-cv & \makecell[l]{Friedman,\\ Nemenyi} \\

    PharmaHGT & MoleculeNet & \makecell[l]{random,\\ scaffold} & 5-cv & no \\

    KANO & MoleculeNet & scaffold & 3-runs & no \\

    GeoDILI & Hepatotoxicity & random & 5-cv & no \\

    MMGIN & TOXRIC,  Tox21 &  random & 1-run & no \\

    3MTox & MoleculeNet &  \makecell[l]{random,\\ scaffold} & 10-runs & no \\

    hERGAT & Cardiotoxicity &  random & 1-run & no \\

    DILI\_GATNN & Hepatotoxicity & random & 10-cv & no \\

    DumplingGNN & MoleculeNet & scaffold & \textit{n}-runs & no \\

    SynthMol & MoleculeNet & \makecell[l]{random,\\ scaffold} & 5-runs & \makecell[l]{Mann\\-Whitney U} \\

    AMPred-LWN & Ames & random & 1-run & no\\

    \textit{ours} & \makecell[l]{TOXRIC,\\Tox21,\\ClinTox,\\Ames} & \makecell[l]{random,\\ scaffold,\\ maxmin,\\ time} & \makecell[l]{5-cv,\\ 5runs} & Plackett–Luce \rule{0pt}{4ex}\\

    \bottomrule
    \end{tabular}
\captionsetup{font=footnotesize}
\caption*{
MoleculeNet includes toxicity datasets such as Tox21, ClinTox, ToxCast, and SIDER, together with non-toxicity molecular benchmarks. TOXRIC provides curated toxicity datasets for multiple organ-specific endpoints, as well as curated versions of datasets such as Tox21, Ames, and ClinTox; here, the term TOXRIC refers specifically to the organ-related toxicity datasets used in our benchmark. \textbf{\textit{Abbreviation:}} cv, cross-validation.
}
\end{table}

\subsection{Experimental set-up}

All experiments were conducted using a standardized experimental protocol to ensure a fair comparison across benchmark datasets and splitting strategies. A 5-fold cross-validation was performed for all partitions where this configuration was applicable. For those splitting strategies that did not support cross-validation —namely \textit{time} and \textit{maxmin} partitions—, models were instead trained and evaluated over five independent runs using different seeds.

The hyperparameter configurations and training setups strictly follow the original implementations described in each model official repository or publication. When a validation set was not originally included, we incorporated one to enable model selection and maintain consistent evaluation conditions. Each model was trained for at least 100 epochs, with an early stopping patience of 20 epochs, even in cases where the original training schedule was shorter.

All experiments were executed on a server equipped with a Dell PowerEdge R750 system featuring 2 × Intel Xeon Gold 6326 processors, 128 GB RAM, and 2 × NVIDIA Ampere A100 80 GB GPUs, running AlmaLinux 8.6 with NVIDIA Driver 515.48.07 (CUDA 11.7). All experiments were implemented in Python, with random seeds fixed for reproducibility.

\subsubsection{Datasets}

For benchmarking purposes, we selected a diverse set of widely used toxicity datasets spanning multiple biological endpoints and prediction tasks. In this work, we employ the clean, curated TOXRIC versions of these datasets \cite{Toxric}, as follows:

\begin{itemize}
\item \textbf{Tox21} is the largest dataset employed in this benchmark, featuring a large-scale in vitro screening collection proposed in the 2014 Tox21 Data Challenge \cite{Tox21} and comprising high-throughput screening data from 12 nuclear receptor and stress response assays to identify potential endocrine disruptors and toxicants via multi-task binary classification.

\item \textbf{ClinTox} specifically targets clinical toxicity, contrasting small molecules that were approved with those that failed in clinical trials due to safety issues, making it particularly valuable in benchmarks that bridge preclinical toxicity modeling and translational risk assessment. The dataset is highly imbalanced, as its collection of molecules originates from the final stages of the drug discovery pipeline —namely clinical trials— where the number of approved compounds significantly outweighs those that fail due to toxicity.

\item \textbf{Ames} dataset captures bacterial mutagenicity outcomes from the Ames test —the canonical benchmark for genotoxicity prediction— often used to assess how well models recover established structure–mutagenicity relationships. It proves to be the simplest task, yielding the best model performance across our benchmark, likely because it stems from a controlled laboratory test where fewer factors contribute to response variability.

\item \textbf{Carcinogenicity} dataset compiles long-term in vivo rodent bioassay outcomes from the Carcinogenic Potency Database (CPDB) labeling compounds by tumor induction potential, enabling evaluation of models' ability to predict oncogenic hazards.

\item \textbf{Hepatotoxicity} dataset aggregates drug-induced liver injury (DILI) annotations from seven authoritative sources, including Liver Toxicity Knowledge Base (LTKB), DILIrank and LiverTox, creating the largest collection for liver-specific toxicity. It provides a critical benchmark for identifying structural patterns linked to hepatic adverse effects, a major cause of drug attrition.

\item \textbf{Cardiotoxicity} datasets are provided at multiple $IC_{50}$ thresholds (1, 5, 10, 30 $\mu M$) to assess hERG channel inhibition —a primary mechanism of QT prolongation and Torsades de Pointes arrhythmia— allowing comprehensive evaluation of model performance across varying clinical risk cutoffs.
\end{itemize}

\begin{figure*}[tb!]
  \centering
  \includegraphics[width=\linewidth]{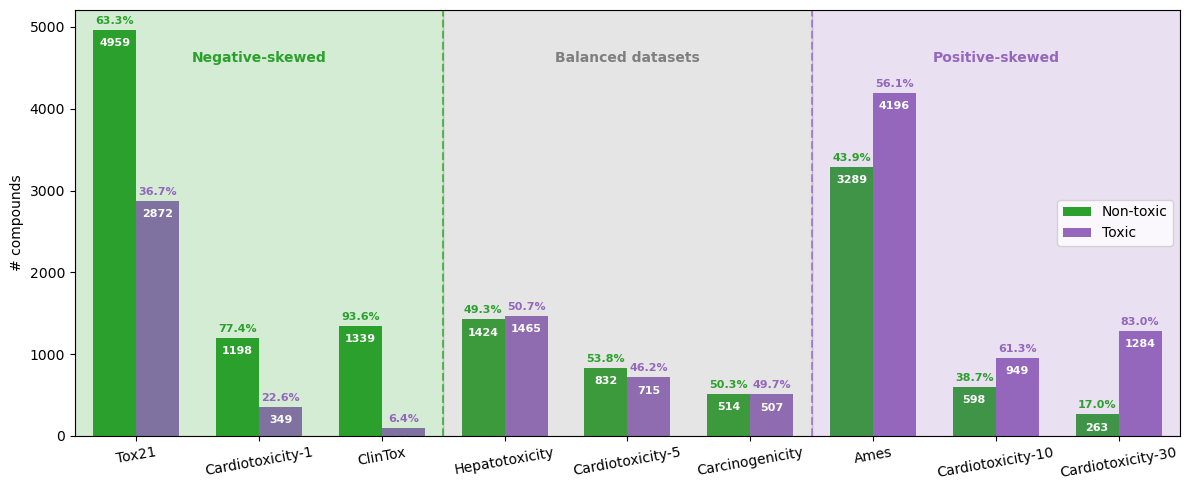}
  \caption{Distribution of positive (toxic) and negative (non-toxic) samples by dataset.}
  \label{img:label-distribution}
\end{figure*}

In \autoref{img:label-distribution} we can see the number of compounds for each collection, as well as the distribution of the positive and negative labels. We can identify three datasets where negative samples dominate, specially in ClinTox, with very few positive samples. Other three datasets are skewed towards the positive label, and the last three are quite well balanced. Cardiotoxicity datasets are spread among all three categories. This is due to the moving threshold that allows for different toxicity consideration of the same molecules.

\subsubsection{Data split}

Different data splitting strategies are crucial in molecular machine learning benchmarks as they simulate real-world deployment scenarios and prevent overly optimistic performance estimates due to data leakage.  Most datasets often overrepresent certain regions of the chemical space, with very few molecular representatives for other less usual chemical motifs. Therefore, random splitting ends up overestimating test performance and undermining generalization. On the other hand, domain-appropriate splits ensure robust evaluation across chemical space, time, and biological contexts. Accordingly, the following splitting strategies were considered:

\begin{itemize}
\item \textbf{Random split} partitions molecules by simple random sampling, preserving overall class balance but allowing structurally similar compounds to appear in both training and test sets by overrepresenting the same chemical class. This strategy suits \textit{i.i.d.} assumptions but fails in chemistry where molecular similarity induces leakage, leading to inflated performance that doesn't generalize to novel chemotypes.

\item \textbf{Scaffold split} groups molecules by their Bemis-Murcko scaffolds \cite{BemisMurcko} —core structures after removing side chains—, ensuring training and test sets contain distinct chemical skeletons. It addresses the similarity leakage of random splits by testing scaffold-hopping —the ability to predict activity for novel core structures critical for virtual screening and lead optimization—. However, although widely adopted, scaffold splitting has been shown to overestimate virtual screening performance by permitting nearly identical scaffolds to be present in both train and test sets \cite{scaffoldsplitsoverestimate}.

\item \textbf{Maxmin split} overcomes this limitation by iteratively selecting test compounds to maximize the minimum distance to any other molecule in the dataset —typically using ECFP or Morgan fingerprints—. This creates the most dissimilar test set possible, rigorously evaluating extrapolation to chemically remote regions of molecular space. This way, the test set has very high coverage of the chemical space and evaluates the generalization of the algorithm to all kinds of compounds in the data \cite{ApisTox}.

\item \textbf{Time split} assigns molecules to splits based on registration and publication dates, ensuring that training data precedes test data chronologically. It mitigates temporal leakage where future knowlege —e.g., recently discovered toxicophores— contaminates training, providing the most realistic estimate for prospective deployment. This split serves as a proxy for extrinsic validation of the models and how they will perform under real-world conditions, rather than the intrinsic validation of theoretical generalizability that the other splitting strategies aim to provide. Therefore, it is essential for retrospective validation of real-world applicability.
\end{itemize}

\textit{Maxmin} and \textit{scaffold} splitting constitute dissimilarity-based approaches, designed to create chemically distinct test sets. In these settings, constructing the validation set using the same dissimilarity criteria can introduce an additional distribution shift between training and validation data, potentially biasing model selection. To mitigate this effect, we adopt a \textit{random validation} %(\textit{randomval}) 
strategy, where the validation set is sampled randomly from the training data distribution. This ensures that hyperparameter tuning and early stopping are performed on a distribution that is representative of the data seen during training, avoiding overfitting to artificially skewed validation splits.

To ensure robust and reliable performance estimation, we adopt different evaluation strategies depending on the nature of the data split. For splitting strategies that allow repeated partitioning —namely \textit{random} and \textit{scaffold} splits— we employ 5-fold cross-validation. In this setting, the dataset is divided into five disjoint folds, and each fold is iteratively used as the test set while the remaining folds are used for training and validation. For splitting strategies where partitioning is fixed by design —such as \textit{maxmin} and \textit{time} splits— cross-validation is not applicable. Instead, models are trained and evaluated over five independent runs using different random seeds, providing an equivalent estimate of variability.

In terms of data allocation, experiments follow a consistent train/validation/test proportion of 60/20/20\% for all splitting strategies with explicit partitioning (\textit{random}, \textit{maxmin}, and \textit{time}). For \textit{scaffold} cross-validation, however, the partitioning follows a different procedure: molecules are first grouped according to their Bemis–Murcko scaffolds, and these scaffold groups are then assigned into five folds. Each fold contains a distinct subset of scaffolds, and evaluation is performed by iteratively using one fold as the test set while training on the remaining scaffold groups, ensuring structural separation between training and test data.

\subsubsection{Metrics}

In binary classification for toxicity prediction, we distinguish correct predictions —True Positives (TP) and True Negatives (TN)— from errors —False Positives (FP), flagging safe compounds as toxic, and False Negatives (FN), missing toxic ones—. 
%In this domain,  \textit{Type II} errors demand greater attention since allowing a genuinely toxic compound to advance risks patient safety and incurs substantial downstream costs in clinical stages. Even so, \textit{Type I} errors also merit consideration, as discarding promising safe candidates prematurely limits therapeutic opportunities that could successfully complete the drug discovery pipeline.
Standard metrics like \textbf{Accuracy} gauge overall correctness but falter on imbalanced datasets by favoring the majority class; \textbf{Recall} tracks detected toxics; \textbf{Precision} assesses positive prediction reliability; and \textbf{Specificity} tracks how many true safe compounds we correctly identified. Yet, each metric tells only part of the story and can be gamed by trivial classifiers.

To address these limitations, the \textbf{F1-score} combines Precision and Recall via their harmonic mean, rewarding models that simultaneously catch toxics reliably while minimizing false alarms. This makes F1 particularly effective in imbalanced settings where no single metric suffices. Building on this, the \textbf{F2-score} further prioritizes Recall to penalize missed toxics more heavily —aligning well with toxicity screening, where undetected risks carry greater consequences than overcaution.

Threshold-independent metrics provide a more comprehensive view by aggregating performance across all classification thresholds. The Area Under the Receiver Operating Characteristic curve (\textbf{AUROC}) summarizes the ROC curve —which plots Sensitivity (\textit{True Positive Rate}) against 1-Specificity (\textit{False Positive Rate})— into a single value measuring a model's ability to rank toxic vs. safe compounds (0.5 = random guessing, 1.0 = perfect discrimination). Unlike threshold-dependent metrics like Precision or Recall, AUROC evaluates the intrinsic ranking quality of predictions by measuring how well the model separates the positive (toxic) and negative (safe) classes across the full spectrum of decision thresholds.

Despite its widespread use, AUROC can be overly optimistic on highly imbalanced datasets typical of toxicity prediction, where abundant negatives make even many False Positives have minimal impact on the False Positive Rate, masking poor toxic detection. The Area Under the Precision-Recall curve (\textbf{AUPR}) addresses this by focusing explicitly on the Precision-Recall trade-off for the positive (toxic) class, offering a more realistic assessment when toxics are rare. Still, AUPR remains sensitive to positive class prevalence and overlooks negative class performance, so both metrics should be considered jointly alongside dataset characteristics and error costs.

For robust evaluation across highly imbalanced toxicity prediction datasets, we adopt the Matthews Correlation Coefficient (\textbf{MCC}) as the primary metric of our benchmark. MCC integrates all four types of results following \autoref{eq:mcc} into a single balanced correlation coefficient ranging from $-1$ (complete disagreement) to $+1$ (perfect prediction), with $0$ corresponding to random performance. Unlike metrics such as Accuracy or AUROC, which may remain artificially high under strong class imbalance, MCC evaluates performance over both positive (toxic) and negative (safe) compounds simultaneously, providing a more reliable estimate of overall classification quality. Similarly, while metrics such as Recall, F2-score, or AUPR mainly emphasize the positive class, MCC balances the impact of the two types of errors regardless of class prevalence.

\begin{equation}
\text{MCC} = \frac{TP \cdot TN - FP \cdot FN}{\sqrt{(TP+FP)(TP+FN)(TN+FP)(TN+FN)}}
\label{eq:mcc}
\end{equation}

Consequently, we use MCC as the main reference metric throughout our benchmark discussion, following several recent toxicity prediction and GNN benchmarking studies that advocate its use under strong class imbalance conditions \cite{GeoDILI, AMPred-LWN}. At the same time, to maintain direct comparability with the broader literature, we also report some other more commonly used evaluation metrics discussed above, all of which can be explored in detail in our online benchmark repository. 
%Under a similar philosophy, \textbf{Cohen’s Kappa} also evaluates performance relative to chance agreement, providing an interpretable estimate of how much better a classifier performs than random guessing. However, given its more limited adoption in toxicity prediction benchmarks, we do not consider it a central metric in our analysis.

\subsubsection{Selected approaches}

Not all approaches included in the review were incorporated into the benchmark. While the review aimed to provide broad coverage of the state-of-the-art approaches, the empirical evaluation required practical constraints to ensure reproducibility and robustness. Therefore, approaches were excluded from the benchmark if they met at least one of the following conditions:

%\paragraph{\textbf{Exclusion criteria}} Works were discarded if they met at least one of the following conditions:
\begin{itemize}
    %\item Studies not fitting any of the considered categories, focusing exclusively on environmental toxicity, ecotoxicology, regulatory toxicology or wet-lab assays without computational modelling.
    %\item Studies that did not include an original model proposal, such as literature reviews or comparative works.
    %\item Studies limited to regression problems, without proposing or evaluating a prediction head for binary classification.
    \item The model was \textbf{restricted to regression problems}, with no binary classification head described or evaluated by the authors. %and authors did not propose or evaluate a prediction head for binary classification.
    %\item Studies without publicly available code.
    \item The source code of the proposed approach was \textbf{not publicly available} in an accesible respository.
    %\item Studies whose code could not be successfully executed following the instructions provided in the repository; the authors were contacted to resolve the issues, but no response was received.
    \item The available code could \textbf{not be successfully executed} following the instructions provided in the repository. In these cases, authors were contacted to resolve the issues, but no response was received.
\end{itemize}

After applying these exclusion criteria, 15 approaches were selected. However, to provide a more complete and exhaustive evaluation, we also included baseline approaches and foundational models in the benchmark, resulting in a total of 20 executed approaches. As baselines, we adopted the implementations of GCN, GAT, and GATGCN presented in \cite{MMGIN}. As foundational models, we selected GROVER and AttentiveFP, two approaches proposed in 2020 that are frequently cited across the selected papers as reference points for comparison.

\section{Results and discussion}
\label{sec:results}

Given the breadth of the benchmark —spanning multiple datasets, partitioning strategies, evaluation metrics, and a total of 21 models— an exhaustive presentation of all possible experimental combinations would be impractical.

We first present an aggregated overview of model performance across datasets, followed by statistical ranking analyses. Then, we motivate the selection of the most informative partitioning strategies and provide a detailed, dataset-level discussion under these settings. The complete set of results, including all metrics, models, datasets, and splitting strategies, is made publicly available through the online benchmark repository\footnote{\url{https://gitlab.citius.gal/noel.suarez/benchtox}}.

\subsection{Global perfomance overview}

\autoref{tab:results_all} summarizes the performance of all evaluated approaches across the selected metrics, aggregated at the dataset level for each partitioning strategy. The best value for each metric and split is highlighted in bold. Additionally, cells are color-coded (green, orange, and yellow) to indicate first, second, and third-ranked approaches in the overall ranking, respectively.

Overall, KPGT consistently ranks among the top-performing methods across most datasets and evaluation settings, getting the first place in 17 out of 28 comparisons over all considered metrics and splits. Other approaches, including 3MTox, GeoDILI, and CD-MVGNN, also demonstrate competitive performance, with a concentration of second and third best places depending on the split and metric considered. Notably, the magnitude of performance differences between these methods is often modest, with less than a 3\% difference for MCC.

%%%%%%%%%%%%%% RESULT TABLE %%%%%%%%%%%%%%%%%%%%%%%%%%%%%%%%%%%%%%%%%%%%%%%%%%%%%%%%
\begin{table*}[p!]
    \centering
    \caption{Benchmark performance comparison across all datasets}
    \label{tab:results_all}
    \small %tiny
    \setlength{\tabcolsep}{6pt}
    \renewcommand{\arraystretch}{0.6}
    %\begin{tabular}{cclccccccc}
    \begin{tabular}{
        >{\hspace{1pt}}c<{\hspace{1pt}}   % Columna 1: más espacio izquierda y derecha
        >{\hspace{8pt}}c<{\hspace{6pt}}   % Columna 2: más espacio izquierda y derecha
        lccccccc  % Resto de columnas igual
    }
    \toprule
    & \textbf{Approach} & \textbf{Split} &
    \textbf{ACC\%} & \textbf{F1\%} & \textbf{F2\%} & \textbf{AUROC\%} & \textbf{AUPR\%} & \textbf{SP\%} & \textbf{MCC\%} \\
    \midrule

    \multirow{12}{*}[-8ex]{\rotatebox[origin=c]{90}{\textbf{2022}}}
    &\multirow{4}{*}[-0.4em]{\textbf{MolCLR}}
        & random & 74.58 & 66.21 & 65.26 & 76.09 & 70.67 & 71.19 & 37.29 \\
        && scaffold & 72.57 & 63.23 & 62.79 & 72.83 & 66.48 & 68.71 & 32.02 \\
        && maxmin & 70.10 & 57.67 & 57.23 & 69.60 & 60.84 & 68.15 & 25.40 \\
        && time & 69.45 & 54.74 & 53.13 & 69.43 & 62.47 & 69.95 & 23.92 \\\cmidrule(lr){2-10}

    &\multirow{4}{*}[-0.4em]{\textbf{GEM}}
        & random & 76.42 & 69.41 & 68.51 & 78.57 & 74.32 & 73.92 & 42.89 \\
        && scaffold & 74.21 & 66.86 & 66.91 & 74.73 & 70.89 & 68.68 & 36.80 \\
        && maxmin & 72.63 & 62.31 & 61.55 & 74.23 & 65.84 & 73.20 & 34.47 \\
        && time & 72.60 & 63.49 & 64.93 & 71.94 & 65.31 & 65.36 & 31.52 \\\cmidrule(lr){2-10}
        
    &\multirow{4}{*}[-0.4em]{\textbf{CD-MVGNN}}
        & random & \cellcolor{ranktwo}77.90 & \cellcolor{rankone}\textbf{71.73} & \cellcolor{ranktwo}71.59 & 80.28 & 76.49 & 70.82 & \cellcolor{ranktwo}44.73 \\
        && scaffold & 74.83 & \cellcolor{rankone}\textbf{68.86} & \cellcolor{ranktwo}69.35 & 77.46 & 73.48 & 67.40 & \cellcolor{rankthree}38.92 \\
        && maxmin & 73.05 & 63.50 & 63.33 & 74.67 & 67.25 & 70.40 & 34.30 \\
        && time & 73.48 & 61.85 & 62.96 & 73.61 & 66.12 & 63.57 & 29.71\\\cmidrule(lr){2-10}

    &\multirow{4}{*}[-0.4em]{\textbf{KPGT}}
        & random & \cellcolor{rankone}\textbf{79.14} & 70.09 & 67.62 & \cellcolor{rankone}\textbf{82.84} & \cellcolor{rankone}\textbf{79.23} & \cellcolor{rankone}\textbf{78.26} & \cellcolor{rankone} \textbf{47.53} \\
        && scaffold & \cellcolor{rankone}\textbf{76.76} & 67.13 & 65.52 & \cellcolor{rankone}\textbf{79.75} & \cellcolor{rankone}\textbf{76.15} & \cellcolor{rankthree}72.84 & \cellcolor{rankone}\textbf{41.58} \\
        && maxmin & \cellcolor{rankone}\textbf{75.74} & \cellcolor{rankthree}64.39 & 60.50 & \cellcolor{rankone}\textbf{78.93} & \cellcolor{rankone}\textbf{73.72} & \cellcolor{rankone}\textbf{78.89} & \cellcolor{rankone}\textbf{41.16} \\
        && time & \cellcolor{rankone}\textbf{75.43} & 59.33 & 55.89 & \cellcolor{rankone}\textbf{78.52} & \cellcolor{rankthree}68.49 & \cellcolor{rankone}\textbf{77.28} & \cellcolor{rankthree}34.33 \\\midrule

    \multirow{12}{*}[-8ex]{\rotatebox[origin=c]{90}{\textbf{2023}}}
    &\multirow{4}{*}[-0.4em]{\textbf{KANO}}
        & random & 76.71 & \cellcolor{ranktwo}70.67 & \cellcolor{rankthree}71.48 & \cellcolor{rankthree}80.67 & \cellcolor{ranktwo}77.44 & 69.21 & 43.44\\
        && scaffold & 74.17 & 65.27 & 63.91 & 77.28 & \cellcolor{ranktwo}74.20 & 70.63 & 37.18\\
        && maxmin & 73.44 & \cellcolor{rankone}\textbf{66.23} & \cellcolor{ranktwo}67.24 & \cellcolor{ranktwo}76.00 & \cellcolor{ranktwo}70.61 & 68.54 & \cellcolor{ranktwo}36.90\\
        && time & 72.67 & 57.76 & 55.35 & 71.25 & 65.09 & 71.68 & 26.54 \\\cmidrule(lr){2-10}

    &\multirow{4}{*}[-0.4em]{\textbf{PharmHGT}}
        & random & 75.86 & 67.93 & 68.24 & 78.14 & 73.22 & 70.08 & 39.72 \\
        && scaffold & 73.61 & 63.42 & 62.75 & 75.18 & 70.03 & 68.94 & 33.71 \\
        && maxmin & 71.92 & 62.09 & 61.91 & 73.63 & 66.28 & 68.57 & 31.79 \\
        && time & 72.47 & 56.38 & 55.00 & 70.19 & 63.90 & 69.54 & 26.26 \\\cmidrule(lr){2-10}

    &\multirow{4}{*}[-0.4em]{\textbf{NYAN}}
        & random & 76.97 & 63.24 & 62.15 & 78.70 & 72.66 & 70.08 & 37.00 \\
        && scaffold & 74.61 & 59.18 & 59.32 & 75.13 & 69.30 & 65.68 & 28.81 \\
        && maxmin & \cellcolor{ranktwo}73.99 & 57.81 & 56.82 & 75.05 & 66.19 & 69.36 & 30.91 \\
        && time & 72.98 & 58.53 & 58.40 & 71.00 & 64.70 & 63.61 & 25.01\\\cmidrule(lr){2-10}

    &\multirow{4}{*}[-0.4em]{\textbf{GeoDILI}}
        & random & 76.66 & 67.79 & 68.08 & 79.48 & 73.80 & 69.98 & 40.10 \\
        && scaffold & 75.24 & 66.37 & 65.98 & \cellcolor{ranktwo}77.96 & 72.80 & 69.66 & 38.56 \\
        && maxmin & 72.97 & 61.45 & 60.78 & \cellcolor{rankthree}75.51 & 68.52 & 70.67 & 34.15 \\
        && time & \cellcolor{rankthree}74.66 & \cellcolor{ranktwo}64.97 & \cellcolor{rankthree}65.92 & \cellcolor{ranktwo}78.43 & \cellcolor{rankone}\textbf{70.69} & 69.10 & \cellcolor{rankone}\textbf{37.38} \\\midrule

    \multirow{12}{*}[1.5ex]{\rotatebox[origin=c]{90}{\textbf{2024}}}
    &\multirow{4}{*}[-0.4em]{\textbf{MMGIN$_{bin}$}}
        & random & 75.77 & 68.79 & 69.07 & 76.64 & 72.10 & 68.46 & 39.58 \\
        && scaffold & 73.82 & 66.49 & 66.99 & 73.62 & 69.22 & 66.41 & 35.43 \\
        && maxmin & 71.78 & 60.29 & 60.39 & 70.49 & 62.98 & 68.35 & 29.72 \\
        && time & 72.43 & 60.63 & 60.87 & 70.68 & 63.04 & 67.49 & 29.10 \\\cmidrule(lr){2-10}
            
    &\multirow{4}{*}[-0.4em]{\textbf{3MTox}}
        & random & \cellcolor{rankthree}77.20 & \cellcolor{rankthree}70.18 & 69.83 & \cellcolor{ranktwo}80.83 & \cellcolor{rankthree}77.24 & 71.61 & \cellcolor{rankthree} 43.64 \\
        && scaffold & \cellcolor{rankthree}75.36 & \cellcolor{ranktwo}68.72 & \cellcolor{rankthree}68.66 & \cellcolor{rankthree}77.72 & \cellcolor{rankthree}73.86 & 68.35 & \cellcolor{ranktwo}39.06 \\
        && maxmin & \cellcolor{rankthree}73.58 & 63.95 & \cellcolor{rankthree}63.43 & 75.43 & \cellcolor{rankthree}69.04 & 69.06 & 34.30\\
        && time & \cellcolor{ranktwo}75.17 & \cellcolor{rankone}\textbf{65.79} & \cellcolor{ranktwo}66.97 & \cellcolor{rankthree}76.40 &  \cellcolor{ranktwo}70.24 & 66.82 & \cellcolor{ranktwo}35.98\\\midrule

    \multirow{12}{*}[-8ex]{\rotatebox[origin=c]{90}{\textbf{2025}}}
    &\multirow{4}{*}[-0.4em]{\textbf{DILI\_GATNN}}
        & random & 77.07 & 61.23 & 60.47 & 80.32 & 76.15 & 71.03 & 40.31 \\
        && scaffold & \cellcolor{ranktwo}75.56 & 60.68 & 59.55 & 77.44 & 73.10 & 70.52 & 32.76 \\
        && maxmin & 73.20 & 55.99 & 55.31 & 75.18 & 68.39 & 70.30 & 32.66 \\
        && time & 74.60 & 57.03 & 55.56 & 75.04 & 66.68 & 70.49 & 29.26 \\\cmidrule(lr){2-10}

    &\multirow{4}{*}[-0.4em]{\textbf{hERGAT}}
        & random & 74.67 & 69.77 & \cellcolor{rankone}\textbf{75.47} & 78.16 & 71.09 & 57.29 & 38.64\\
        && scaffold & 71.57 & \cellcolor{rankthree}67.98 & \cellcolor{rankone}\textbf{74.84} & 74.61 & 67.96 & 50.19 & 32.65 \\
        && maxmin & 70.23 & \cellcolor{ranktwo}64.85 & \cellcolor{rankone}\textbf{71.63} & 73.02 & 63.40 & 50.70 & 28.84\\
        && time & 69.70 & \cellcolor{rankthree}64.69 & \cellcolor{rankone}\textbf{71.72} & 70.38 & \cellcolor{rankthree}64.51 & 43.31 & 26.27 \\\cmidrule(lr){2-10}

    &\multirow{4}{*}[-0.4em]{\textbf{DumplingGNN}}
        & random & 70.61 & 54.72 & 55.11 & 66.39 & 59.68 & 61.54 & 18.49 \\
        && scaffold & 70.43 & 56.22 & 56.51 & 67.12 & 61.05 & 61.60 & 21.65 \\
        && maxmin & 68.49 & 50.30 & 51.25 & 65.07 & 53.33 & 63.43 & 16.56 \\
        && time & 67.97 & 51.45 & 51.21 & 65.75 & 58.35 & 61.72 & 15.07 \\\cmidrule(lr){2-10}

    &\multirow{4}{*}[-0.4em]{\textbf{SynthMol}}
        & random & 75.09 & 66.15 & 65.84 & 78.19 & 74.32 & 72.70 & 38.96 \\
        && scaffold & 73.89 & 65.03 & 64.04 & 75.78 & 71.43 & 71.27 & 35.95 \\
        && maxmin & 72.31 & 61.23 & 60.12 & 74.76 & 67.81 & \cellcolor{rankthree}75.20 & \cellcolor{rankthree}34.53\\
        && time & 70.65 & 58.20 & 58.65 & 70.61 & 63.33 & 63.90 & 25.58\\\midrule

    \multirow{12}{*}[2.8em]{\rotatebox[origin=c]{90}{\textbf{2026}}}
&\multirow{4}{*}[-0.4em]{\textbf{AMPred-LWN}}
       & random & 70.34 & 62.49 & 62.92 & 79.37 & 74.48 & 77.54 & 39.68 \\
       && scaffold & 68.26 & 60.80 & 61.63 & 76.47 & 71.75 & 72.10 & 34.14 \\
       && maxmin & 67.88 & 57.93 & 58.46 & 73.72 & 65.84 & 74.01 & 32.08\\
       && time & 65.14 & 58.58 & 60.76 & 74.22 & 64.40 & 62.00 & 27.52\\\bottomrule

    \end{tabular}
\captionsetup{font=footnotesize}
\caption*{
\centering
\parbox{0.9\linewidth}{\centering
Best metric for each split in \textbf{bold}. Green, orange and yellow highlights indicate
the first, second and third-best approach \\in the general ranking (continues in next table).
}
}
\end{table*}
%%%%%%%%%%%%%%%%%%%%%%%%%%%%%%%%%%%%%%%%%%%%%%%%%%%%%%%%%%%%%%%%%%%%%%%%%%%%%%%%%%%%

Within this general pattern, hERGAT stands out primarily in the metrics that emphasize positive-class recovery, achieving the best results in F2-score with a 71-75\% and also performing strongly in F1-score, where it frequently ranks third or second.  However, this advantage does not extend to the remaining metrics, suggesting a classification behavior that prioritizes sensitivity toward toxic compounds rather than uniformly balanced performance across all criteria. In practical terms, hERGAT may be particularly suitable for safety-oriented screening scenarios, where minimizing false negatives is more important than maintaining a strict balance between precision and recall.

Regarding the comparison with the baseline and foundation models reported in \autoref{tab:results_baselines}, simple GNN architectures —such as GCN and GAT— as well as widely used reference models —namely AttentiveFP and GROVER— achieve results that are frequently close to those of more recent approaches, with around a 2\% difference in MCC with most approaches and up to a 4\% with those just bellow the third-best. This suggests that most of the proposed methods —excluding KPGT, CD-MVGNN, 3MTox, and GeoDILI— provide only limited improvements over established baselines.

%%%%%%%%%%%%%% BASELINE TABLE %%%%%%%%%%%%%%%%%%%%%%%%%%%%%%%%%%%%%%%%%%%%%%%%%%%%%%%%
\begin{table*}[tb!]
    \centering
    \caption{Performance comparison of baseline methods across all datasets}
    \label{tab:results_baselines}
    \small %tiny
    \setlength{\tabcolsep}{6pt}
    \renewcommand{\arraystretch}{0.6}
    %\begin{tabular}{cclccccccc}
    \begin{tabular}{
        >{\hspace{1pt}}c<{\hspace{1pt}}   % Columna 1: más espacio izquierda y derecha
        >{\hspace{8pt}}c<{\hspace{6pt}}   % Columna 2: más espacio izquierda y derecha
        lccccccc  % Resto de columnas igual
    }
    \toprule
    & \textbf{Approach} & \textbf{Split} &
    \textbf{ACC\%} & \textbf{F1\%} & \textbf{F2\%} & \textbf{AUROC\%} & \textbf{AUPR\%} & \textbf{SP\%} & \textbf{MCC\%} \\
    \midrule

    \multirow{12}{*}[-3ex]{\rotatebox[origin=c]{90}{baseline}}
    & \multirow{4}{*}[-0.4em]{\textbf{GCN}}
        & random & 76.28 & 67.87 & 67.99 & \textbf{77.34} & \textbf{71.28} & 70.36 & 40.03 \\
        && scaffold & 73.69 & \textbf{64.08} & 65.02 & \textbf{74.41} & 67.79 & 66.96 & 33.75 \\
        && maxmin & 71.73 & 61.25 & \textbf{61.86} & 72.70 & 64.28 & 69.24 & 31.74 \\
        && time & \textbf{72.07} & 61.98 & 63.28 & \textbf{72.84} & 63.52 & \textbf{65.44} & \textbf{30.31}\\\cmidrule(lr){2-10}

    & \multirow{4}{*}[-0.4em]{\textbf{GAT}}
        & random & \textbf{76.26} & 67.87 & 67.88 & 77.21 & 71.05 & \textbf{70.72 }& \textbf{40.20} \\
        && scaffold & 73.28 & 63.92 & \textbf{65.11} & 74.09 & \textbf{67.81} & 65.88 & 32.94 \\
        && maxmin & 71.97 & 60.60 & 60.65 & 72.36 & 63.84 & 70.44 & 31.42 \\
        && time & 71.81 & \textbf{62.25} & \textbf{64.44} & 72.75 & 63.41 & 62.99 & 29.44 \\\cmidrule(lr){2-10}

    & \multirow{4}{*}[-0.4em]{\textbf{GATGCN}}
        & random & 76.12 & \textbf{68.07} & \textbf{68.72} & 77.32 & 70.99 & 69.89 & 39.92 \\
        && scaffold & \textbf{73.73} & 63.94 & 64.45 & 73.97 & 67.23 & \textbf{67.49} & \textbf{33.98} \\
        && maxmin & \textbf{71.99} & \textbf{61.83} & 61.69 & \textbf{72.80} & \textbf{64.34} & \textbf{70.54} & \textbf{32.67} \\
        && time & 71.89 & 61.87 & 63.28 & 72.78 & \textbf{63.81} & 64.65 & 29.58 \\\midrule\\\midrule

    \multirow{12}{*}[-3ex]{\rotatebox[origin=c]{90}{foundational  (\textbf{2020})}}
    &\multirow{4}{*}[-0.4em]{\textbf{GROVER$_{base}$}}
        & random & \textbf{75.19} & 55.77 & 52.82 & 76.27 & 70.63 & \cellcolor{ranktwo}\textbf{75.54} & 31.29 \\
        && scaffold & \textbf{73.96} & 54.93 & 52.61 & \textbf{75.05} & \textbf{69.41} & \cellcolor{rankone}\textbf{73.79} & 29.16 \\
        && maxmin & \textbf{71.74} & 47.36 & 44.62 & \textbf{71.35} & 62.06 & \cellcolor{ranktwo}\textbf{75.95} & 22.38 \\
        && time & 71.36 & \textbf{51.02} & \textbf{47.63} & \textbf{69.07} & 62.36 & \cellcolor{rankthree}74.61 & \textbf{24.49} \\\cmidrule(lr){2-10}
        
    & \multirow{4}{*}[-0.4em]{\textbf{GROVER$_{large}$}}
        & random & 73.55 & 55.83 & 48.53 & 75.20 & 68.92 &  \cellcolor{rankthree}75.14 & 27.04 \\
        && scaffold & 73.05 & 51.54 & 49.37 & 74.02 & 67.29 & \cellcolor{ranktwo}73.56 & 25.97 \\
        && maxmin & 71.65 & 46.94 & 44.39 & 70.88 & 61.10 & 75.07 & 20.77 \\
        && time & \textbf{71.52} & 49.81 & 46.71 & 68.03 & \textbf{62.68} & 74.59 & 23.17\\\cmidrule(lr){2-10}
        
    & \multirow{4}{*}[-0.4em]{\textbf{AttentiveFP}}
        & random & 70.79 & \textbf{63.46} & \textbf{65.04} & \textbf{77.08} & \textbf{71.60} & 71.52 & \textbf{36.51} \\
        && scaffold & 68.65 & \textbf{59.94} &\textbf{ 61.99} & 73.72 & 68.77 & 67.43 & \textbf{30.61} \\
        && maxmin & 65.90 & \textbf{58.07} & \textbf{59.87} & 71.26 & \textbf{63.56} & 69.83 & \textbf{29.87} \\
        && time & 62.58 & 47.26 & 44.68 & 68.15 & 61.46 & \cellcolor{ranktwo}\textbf{75.91} & 21.01
        \\\bottomrule
    
    \end{tabular}
\captionsetup{font=footnotesize}
\caption*{
Best metric for each split and table division is hightlighted in \textbf{bold}. Colour rankings continue those of previous table.
}
\end{table*}
%%%%%%%%%%%%%%%%%%%%%%%%%%%%%%%%%%%%%%%%%%%%%%%%%%%%%%%%%%%%%%%%%%%%%%%%%%%%%%%%%%%%

In particular, foundational approaches like GROVER and AttentiveFP stand out for their comparatively strong specificity, indicating that they are especially effective at correctly identifying non-toxic compounds. This represents a relevant advantage in toxicity screening, where reducing false positives helps avoid the premature discard of safe and potentially useful candidates. However, this behavior also suggests that these models may achieve their specificity gains by sacrificing sensitivity, that is, by being more conservative in assigning compounds to the toxic class. In contrast, several more recent approaches appear to shift this trade-off toward higher toxic-compound recovery, often at the expense of a lower ability to reject safe molecules. From this perspective, the newer models are not necessarily outperforming the 2020 baselines across the board, but rather redistributing their errors toward a different balance between the two classes.

\subsection{Statistical ranking across experiments}

To enable a robust comparison across heterogeneous evaluation settings, we convert the MCC values obtained for each combination of dataset and splitting strategy into model rankings and aggregate them using the Plackett–Luce model. The model estimates the relative worth of each approach, which, after normalization, can be interpreted as its probability of ranking first among the compared methods. By relying on rankings rather than raw metric values, the analysis is less sensitive to differences in scale across datasets and enables statistically significant differences between methods to be identified while accounting for variability across experimental settings.

\begin{figure*}[ht!]
  \centering
  \includegraphics[width=\linewidth]{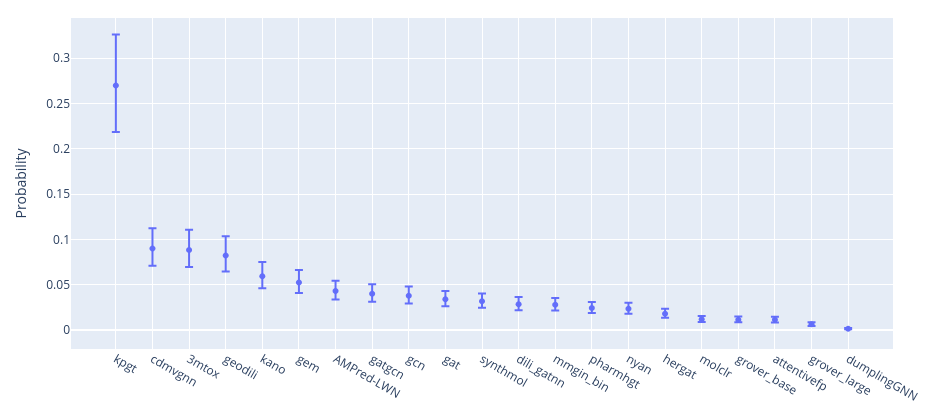}
  \caption{Plackett–Luce ranking for MCC over all experiments.}
  \label{img:ranking_all_mcc}
\end{figure*}

As shown in \autoref{img:ranking_all_mcc}, the resulting rankings reveal a clear hierarchy among the evaluated models. KPGT occupies the top position with an estimated probability of ranking first of approximately 0.27, placing it well above the remaining approaches. The next-best models lie below 0.10, and their confidence intervals do not overlap with that of KPGT, indicating that the observed advantage is not only consistent but also statistically robust across the different experimental configurations.

Below this top performer, a second separable tier is formed by CD-MVGNN, 3MTox, and GeoDILI. These three methods yield practically indistinguishable performances, as their point estimates cluster tightly together and their confidence intervals overlap extensively, indicating that no meaningful statistical differences exist among them. When contrasted with the remaining approaches placed further down the ranking, this group demonstrates clear and statistically significant superiority. However, the lower boundary of this tier partially overlaps with the confidence intervals of KANO and GEM, which are positioned slightly lower in the aggregated ranking. This prevents us from unequivocally asserting that CD-MVGNN, 3MTox, and GeoDILI are strictly better than KANO and GEM in every single experimental scenario, even though they do consistently surpass all methods positioned below this overlapping pair. %cluster

\subsection{Influence of partitioning strategy}

\begin{figure}[htb!]
  \centering
  \includegraphics[width=\linewidth]{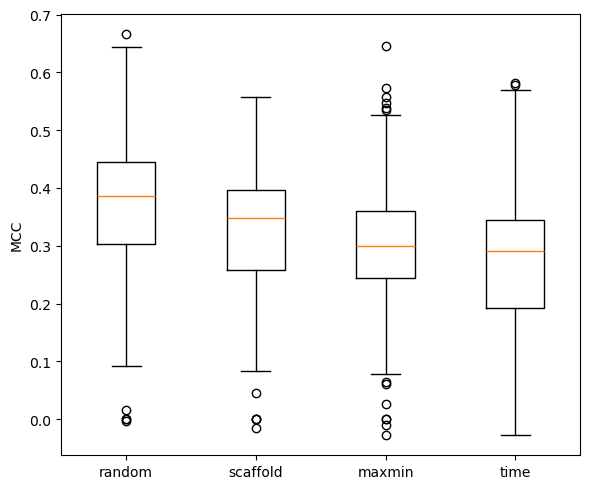}
  \caption{Boxplot comparing results obtained by different approaches under each partition strategy.}
  \label{img:boxplot}
\end{figure}

The data splitting strategy strongly influences both model performance and model rankings. As shown in the boxplot in \autoref{img:boxplot}, both \textit{maxmin} and \textit{time} splits stand out as the most challenging partitioning strategies, yielding average MCC differences of approximately 10\% relative to the \textit{random} split. This difference may indicate that evaluations based on \textit{random} or \textit{scaffold}-based splits sample from overlapping regions of the chemical space during both training and testing, allowing a degree of structural similarity leakage \cite{scaffoldsplitsoverestimate, ApisTox}. In contrast, \textit{maxmin} and \textit{time}-based splits provide evaluations of model generalisation to structurally dissimilar and chronologically later compounds, respectively.

Specifically, the \textit{maxmin} split enforces maximal structural dissimilarity between training and test instances, challenging models to generalize beyond the molecular chemotypes encountered during training, while the \textit{time} split imposes a strict chronological partition based on the year of compound publication, thereby simulating prospective evaluation and offering a closer approximation of how these tools would perform in actual screening workflows. Since these scenarios reflect real-world deployment, we focus the remainder of our comparative analysis on them.

\begin{figure*}[tb!]
  \centering
  \includegraphics[width=\linewidth]{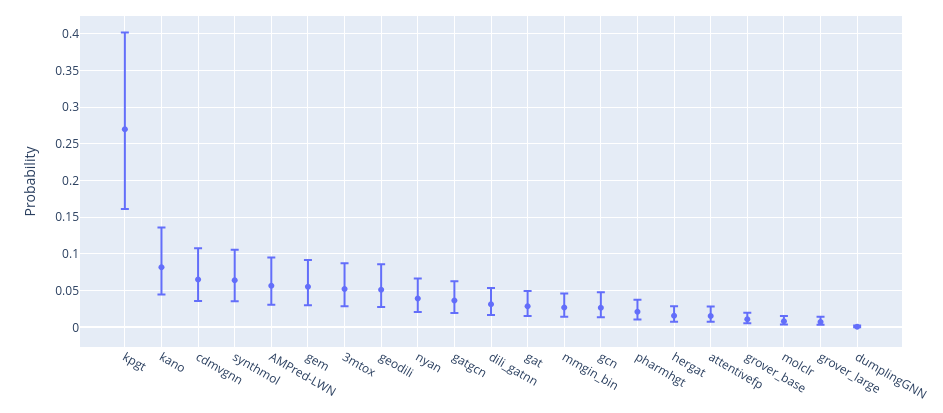} 
  \caption{Plackett–Luce ranking for MCC over \textit{maxmin} split.}
  \label{img:maxmin_mcc}
\end{figure*}

As shown in \autoref{img:maxmin_mcc}, KPGT attains the highest performance under the \textit{maxmin} split, with a statistically significant margin over all other methods, as its confidence interval does not overlap with any competitor. The remaining models cluster within a narrow performance range, with KANO ranking second. However, the partial overlap between KANO and baseline GNNs indicates that its advantage over foundational architectures is limited.

The performance of KPGT and KANO may be explained by their respective pretraining strategies, designed to learn molecular representations enriched with chemical domain knowledge. KPGT incorporates expert-designed molecular fingerprints into a self-supervised learning framework, in which the model is trained to reconstruct the masked input from the available molecular information. KANO, in turn, exploits a knowledge graph (\textit{ElementKG}) and functional prompts to integrate knowledge derived from periodic-table properties. It further employs a contrastive learning strategy that dynamically enriches molecular inputs with information about known functional groups.

\begin{figure*}[tb!]
  \centering
  \includegraphics[width=\linewidth]{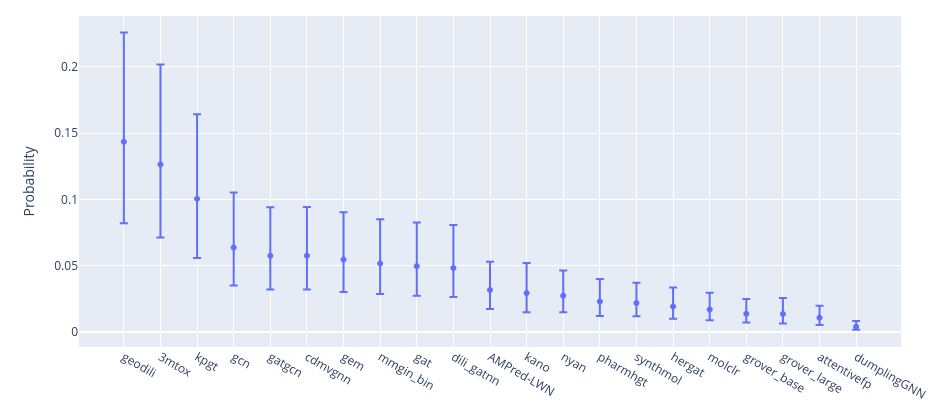}
  \caption{Plackett–Luce ranking for MCC over \textit{time} split.}
  \label{img:time_mcc}
\end{figure*}

On the \textit{time} split, the performance ranking shifts substantially.
In \autoref{img:time_mcc}, GeoDILI and 3MTox attain the highest MCC values, while KPGT falls from first to third place relative to its top-ranked performance under the \textit{maxmin} split. However, the confidence intervals of the three methods largely overlap under the temporal split; therefore, no clear statistical ranking can be established among them. In fact, even though KPGT remains within the upper tier, its confidence interval overlaps considerably with those of several other models, including foundational GNNs such as GCN, GAT or GATGCN. % This contrast reinforces the earlier observation that model rankings are highly contingent upon the evaluation protocol, underscoring the necessity of employing multiple, realistically challenging partitioning strategies to draw robust conclusions about model generalizability.

A plausible explanation for this result lies in the nature of the knowledge KPGT and KANO encode. Both approaches are strongly oriented toward capturing structural dissimilarity and leveraging predefined chemical knowledge or patterns derived from historical data. However, in time split, recently introduced molecules may not conform to traditional structural patterns or well-characterized functional groups embedded in these models. Consequently, methods such as GeoDILI and 3MTox become more competitive: GeoDILI benefits from incorporating three-dimensional geometric information, particularly bond-angle relationships, while 3MTox leverages motif-based representations —e.g., ring structures— that provide greater flexibility in identifying emerging functional patterns, unlike KANO, which relies on expert-curated functional-group definitions. Nevertheless, KPGT retains a strong position even in this split, which may be attributed to its bond-centric representation and its fingerprint-guided pretraining strategy. This combination embeds rich chemical knowledge and may also enable the model to uncover previously unseen toxic or functional patterns, partially mitigating the limitations associated with temporal extrapolation.

Therefore, model rankings are highly dependent on the evaluation split. While knowledge-enhanced approaches appear to be particularly competitive under the \textit{maxmin} split, no clear advantage emerges under the \textit{time} split. Instead, a diverse set of methods achieve comparable performance. These results highlight the importance of evaluating toxicity prediction models across multiple partitioning strategies, as conclusions drawn from a single experimental setting may not generalize to more realistic deployment scenarios.

\subsection{Dataset-level analysis under challenging splits}

We now provide a detailed analysis of model performance at the dataset level under the \textit{maxmin} and \textit{time} partitioning strategies, focusing on the five top-performing approaches shown %identified through the statistical ranking 
in \autoref{img:ranking_all_mcc}: KPGT, CD-MVGNN, 3MTox, GeoDILI, and KANO.

Under the \textit{maxmin} split (\autoref{tab:maxmin_dataset}), KPGT and KANO emerge as the most consistently competitive methods across datasets.
%, although their relative advantage is not uniform
KPGT attains the best average performance in all metrics, with particularly strong gains in AUROC and MCC, reaching 78.93 and 41.16, respectively; while KANO follows closely with 76.00 in AUROC and 36.90 in MCC. At dataset level, %KPGT is the top-performing model on 8 of the 9 datasets for AUROC and on 7 of the 9 for MCC
KPGT achieves the highest AUROC on 8 of the 9 datasets and the highest MCC on 7 of the 9, indicating that its performance advantage is not driven by a single scenario %but rather by broad robustness 
but is consistently observed across chemically diverse toxicological datasets. The largest gaps in MCC are observed in ClinTox, where KPGT %(55.80)
exceeds CD-MVGNN 
%(49.49)
and GeoDILI
%v(44.80) 
by 6.31 and 11.00 points, respectively; and in Ames, where it improves over KANO 
% (51.61) 
and GeoDILI 
% (51.24) 
by 5.65 and 6.02 points, respectively. These results suggest that KPGT benefits from representations that generalize particularly well under strict chemical extrapolation, likely due to its large-scale pretraining and stronger capture of latent molecular structure. By contrast, KANO appears competitive on ClinTox, where it clearly achieves the highest AUROC (93.49) and MCC (64.53), showing that ontology-informed features can be advantageous in data-scarce scenarios where biological or dataset-specific prior knowledge provides valuable guidance.

\begin{table*}[tb!]
    \centering
    \caption{Maxmin-split performance by dataset}
    \label{tab:maxmin_dataset}
    \small %tiny
    \setlength{\tabcolsep}{2pt}
    \renewcommand{\arraystretch}{0.5}
    \begin{tabular}{clcccccccccc}
    \toprule
    \textbf{Approach} & \textbf{Metric} &
    \textbf{Tox21} & \textbf{ClinTox} & \textbf{Ames} & \textbf{Carcino} & \textbf{Cardio-1} & \textbf{Cardio-5} & \textbf{Cardio-10} & \textbf{Cardio-30} & \textbf{Hepato} & \textbf{\textit{AVG.}} \\
    \midrule

    \multirow{3}{*}{\textbf{CD-MVGNN}}
        & auroc & 72.43 & 89.73& 82.63& \cellcolor{rankthree}69.60 & \cellcolor{ranktwo}81.39 & 74.00 & \cellcolor{rankthree}67.92 & 66.00 & 68.33 & \textit{74.67} \\
        & aupr & 62.78 & 51.28 & 78.20 & 65.53 & \cellcolor{ranktwo}56.67 & 64.34 & 69.44 & 89.12 & 67.88 & \textit{67.25} \\
        & mcc & \cellcolor{rankthree}35.18 & 49.49 & 50.78 & 25.21 & \cellcolor{rankthree}40.50 & 34.48 & \cellcolor{ranktwo}30.15 & 12.91 & \cellcolor{rankthree}30.03 & \cellcolor{rankthree}\textit{34.30} \\\midrule

    \multirow{3}{*}{\textbf{KPGT}}
        & auroc & \cellcolor{rankone}76.13 & \cellcolor{ranktwo}92.17 & \cellcolor{rankone}86.35 & \cellcolor{rankone}\textbf{72.82} & \cellcolor{rankone}\textbf{84.82} & \cellcolor{rankone}76.97 & \cellcolor{rankone}72.43 & \cellcolor{rankone}\textbf{75.84} & \cellcolor{rankone}72.81 & \cellcolor{rankone}\textit{\textbf{78.93}} \\
        & aupr & \cellcolor{rankone}\textbf{67.39} & \cellcolor{ranktwo}66.81 & \cellcolor{rankone}83.96 & \cellcolor{rankone}70.54 & \cellcolor{rankone}61.02 & \cellcolor{rankone}68.99 & \cellcolor{rankone}75.59 & \cellcolor{rankone}\textbf{92.99} & \cellcolor{rankone}\textbf{76.21} & \cellcolor{rankone}\textit{\textbf{73.72}}\\
        & mcc & \cellcolor{rankone}\textbf{38.88} & \cellcolor{ranktwo}55.80 & \cellcolor{rankone}57.26 & \cellcolor{rankone}\textbf{33.76} & \cellcolor{rankone}49.95 & \cellcolor{rankone}40.31 & \cellcolor{rankone}32.96 & \cellcolor{ranktwo}29.56 & \cellcolor{rankone}31.93 & \cellcolor{rankone}\textit{\textbf{41.16}}\\\midrule

    \multirow{3}{*}{\textbf{KANO}}
        & auroc & \cellcolor{rankthree}73.66 & \cellcolor{rankone}93.49 & 82.52 & \cellcolor{ranktwo}70.73 & \cellcolor{rankthree}80.71 & \cellcolor{ranktwo}75.89 & \cellcolor{ranktwo}67.99 & 68.26 & \cellcolor{rankthree}70.73 & \cellcolor{ranktwo}\textit{76.00} \\
        & aupr & \cellcolor{ranktwo}65.07 & \cellcolor{rankone}\textbf{72.24} & 78.21 &\cellcolor{ranktwo} 68.07 & \cellcolor{rankthree}53.28 & \cellcolor{ranktwo}66.31 & \cellcolor{rankthree}70.70 & 89.39 & 72.20 & \cellcolor{ranktwo}\textit{70.61} \\
        & mcc & 34.90 & \cellcolor{rankone}\textbf{64.53} & \cellcolor{ranktwo}51.61 & \cellcolor{ranktwo}33.63 & 38.93 & 37.26 & \cellcolor{rankthree}27.57 & 12.74 & \cellcolor{ranktwo}30.90 & \cellcolor{ranktwo}\textit{36.90} \\\midrule

    \multirow{3}{*}{\textbf{GeoDILI}}
        & auroc & 72.76 & 87.73 & \cellcolor{rankthree}82.65 & 67.59 & 80.27 & 74.33 & 67.19 & \cellcolor{ranktwo}75.36 & \cellcolor{ranktwo}71.70 & \cellcolor{rankthree}\textit{75.51} \\
        & aupr & 62.82 & 56.87 & \cellcolor{rankthree}78.44 & 63.62 & 51.86 & 64.34 & \cellcolor{ranktwo}70.76 & \cellcolor{ranktwo}92.32 & \cellcolor{ranktwo}75.67 & \textit{68.52} \\
        & mcc & 34.83 & 44.80 & \cellcolor{rankthree}51.24 & \cellcolor{rankthree}25.42 & 30.25 & \cellcolor{rankthree}37.28 & 21.51 & \cellcolor{rankone}\textbf{32.01} & 30.02 & \textit{34.15} \\\midrule

    \multirow{3}{*}{\textbf{3MTox}}
        & auroc & \cellcolor{ranktwo}74.10 & \cellcolor{rankthree}91.90 & \cellcolor{ranktwo}82.99 & 67.67 & 78.84 & \cellcolor{rankthree}75.36 & 67.08 & \cellcolor{rankthree}71.76 & 69.17 & \textit{75.43} \\
        & aupr & \cellcolor{rankthree}65.03 & \cellcolor{rankthree}62.48 & \cellcolor{ranktwo}79.31 & \cellcolor{rankthree}65.58 & 50.82 & \cellcolor{rankthree}64.42 & 69.34 & \cellcolor{rankthree}91.83 & \cellcolor{rankthree}72.54 & \cellcolor{rankthree}\textit{69.04} \\
        & mcc & \cellcolor{ranktwo}35.96 & \cellcolor{rankthree}53.48 & 50.29 & 25.30 & \cellcolor{ranktwo}40.76 & \cellcolor{ranktwo}38.76 & 22.99 & \cellcolor{rankthree}13.94 & 27.22 & \textit{34.30} \\\bottomrule

    \end{tabular}
\end{table*}

Regarding the \textit{time} split (\autoref{tab:time_dataset}), %the relative ranking among the top-performing methods becomes less stable and more dataset-dependent.
GeoDILI, 3MTox, and KPGT occupy the top tier, with each method achieving first place in 3 out of the 9 datasets, showing similar performance across datasets, with no single method clearly dominating. This similarity is also reflected in their average MCC values after excluding ClinTox: 38.77 for KPGT, 38.14 for GeoDILI, and 36.26 for 3MTox.

The inclusion of the ClinTox dataset, however, has a major impact on the overall ranking. In this dataset, clear differences between methods are observed across all metrics, especially in MCC, which falls below zero for KPGT. While AUROC remains relatively high for all methods, the lower AUPR and MCC values of KPGT indicate that it struggles to identify the minority toxic class under strong class imbalance. In contrast, the higher AUPR values for GeoDILI and 3MTox indicate a better trade-off between precision and recall. This is also reflected in their positive MCC scores and more balanced classification performance.

Across the remaining datasets, the differences in MCC among the five approaches vary considerably. On Tox21, all methods obtain relatively similar results, with a difference of only 3.38 points between the highest and lowest MCC values. The gap is also moderate on Carcino (5.03) and Cardio-10 (6.22), but is larger on Ames (9.60), Cardio-5 (10.74), and Hepato (8.33). The largest differences are observed on Cardio-1 (21.60) and Cardio-30 (33.20). Overall, these results show that the relative performance of the approaches under the \textit{time} split depends strongly on the specific characteristics of each dataset.

\begin{table*}[tb!]
    \centering
    \caption{Time-split performance by dataset}
    \label{tab:time_dataset}
    \small %tiny
    \setlength{\tabcolsep}{2pt}
    \renewcommand{\arraystretch}{0.5}
    \begin{tabular}{clcccccccccc}
    \toprule
    \textbf{Approach} & \textbf{Metric} &
    \textbf{Tox21} & \textbf{ClinTox} & \textbf{Ames} & \textbf{Carcino} & \textbf{Cardio-1} & \textbf{Cardio-5} & \textbf{Cardio-10} & \textbf{Cardio-30} & \textbf{Hepato} & \textbf{\textit{AVG.}} \\
    \midrule

    \multirow{3}{*}{\textbf{CD-MVGNN}}
        & auroc & 73.55 & 79.23 & 83.89 & \cellcolor{rankthree}70.30 & 72.58 & 71.69 & 72.13 & 64.01 & \cellcolor{rankthree}75.15 & \textit{73.61} \\
        & aupr & 61.98 & \cellcolor{rankthree}15.04 & 88.72 & \cellcolor{ranktwo}74.46 & 52.22 & 66.36 & \cellcolor{rankthree}78.63 & 87.31 & \cellcolor{rankthree}70.39 & \textit{66.12} \\
        & mcc & \cellcolor{rankthree}35.91 & \cellcolor{rankthree}10.46 & 51.90 & \cellcolor{ranktwo}30.41 & 28.43 & 30.56 & 31.15 & 14.18 & \cellcolor{rankthree}34.42 & \textit{29.71} \\\midrule

    \multirow{3}{*}{\textbf{KPGT}}
        & auroc & \cellcolor{rankone}\textbf{76.22} & \cellcolor{rankthree}90.60 & \cellcolor{rankone}\textbf{87.88} & \cellcolor{rankone}71.40 & \cellcolor{rankone}82.58 & \cellcolor{rankthree}72.86 & \cellcolor{ranktwo}72.88 & \cellcolor{rankone}74.33 & \cellcolor{rankone}\textbf{77.96} & \cellcolor{rankone}\textit{78.52} \\
        & aupr & \cellcolor{rankone}65.87 & 6.70 & \cellcolor{rankone}\textbf{92.26} & \cellcolor{rankone}\textbf{76.23} & \cellcolor{rankone}\textbf{61.17} & \cellcolor{rankthree}69.29 & 78.57 & \cellcolor{rankone}91.98 & \cellcolor{rankone}74.33 & \cellcolor{rankthree}\textit{68.49} \\
        & mcc & 35.82 & -1.21 & \cellcolor{rankone}\textbf{57.77} & \cellcolor{rankthree}30.37 & \cellcolor{rankone}\textbf{50.03} & \cellcolor{rankthree}34.77 & \cellcolor{ranktwo}35.31 & \cellcolor{rankone}30.59 & \cellcolor{ranktwo}35.52 & \cellcolor{rankthree}\textit{34.33} \\\midrule

    \multirow{3}{*}{\textbf{KANO}}
        & auroc & 74.51 & 66.76 & 83.81 & 67.91 & 68.47 & 71.61 & \cellcolor{rankthree}72.66 & 62.47 & 73.06 & \textit{71.25} \\
        & aupr & 63.67 & 2.95 & \cellcolor{ranktwo}89.50 & 73.08 & 51.17 & 69.20 & \cellcolor{ranktwo}79.57 & \cellcolor{rankthree}87.86 & 68.79 & \textit{65.09} \\
        & mcc & 34.14 & -1.37 & 48.17 & 28.21 & 33.47 & 32.10 & 33.98 & -2.61 & 32.80 & \textit{26.54} \\\midrule

    \multirow{3}{*}{\textbf{GeoDILI}}
        & auroc & \cellcolor{rankthree}74.58 & \cellcolor{rankone}\textbf{98.59} & \cellcolor{rankthree}84.45 & 65.09 & \cellcolor{ranktwo}79.32 & \cellcolor{ranktwo}77.02 & \cellcolor{rankone}\textbf{77.19} & \cellcolor{ranktwo}72.36 & \cellcolor{ranktwo}77.24 & \cellcolor{ranktwo}\textit{78.43} \\
        & aupr & \cellcolor{rankthree}63.85 & \cellcolor{ranktwo}37.01 & 88.38 & 69.44 & \cellcolor{ranktwo}56.31 & \cellcolor{rankone}\textbf{71.74} & \cellcolor{rankone}\textbf{84.68} & \cellcolor{ranktwo}90.64 & \cellcolor{ranktwo}74.15 & \cellcolor{rankone}\textit{70.69} \\
        & mcc & \cellcolor{ranktwo}36.43 & \cellcolor{ranktwo}31.28 & \cellcolor{rankthree}53.91 & 26.85 & \cellcolor{ranktwo}42.32 & \cellcolor{rankone}\textbf{41.30} & \cellcolor{rankone}\textbf{37.37} & \cellcolor{ranktwo}27.96 & \cellcolor{rankone}\textbf{38.96} & \cellcolor{rankone}\textit{37.38} \\\midrule

    \multirow{3}{*}{\textbf{3MTox}}
        & auroc & \cellcolor{ranktwo}76.02 & \cellcolor{ranktwo}92.08 & \cellcolor{ranktwo}84.56 & \cellcolor{ranktwo}70.42 & \cellcolor{rankthree}75.93 & \cellcolor{rankone}\textbf{77.21} & 70.62 & \cellcolor{rankthree}67.17 & 73.64 & \cellcolor{rankthree}\textit{76.40} \\
        & aupr & \cellcolor{ranktwo}65.16 & \cellcolor{rankone}47.28 & \cellcolor{rankthree}89.08 & \cellcolor{rankthree}73.17 & \cellcolor{rankthree}52.52 & \cellcolor{ranktwo}70.74 & 75.76 & 88.73 & 69.73 & \cellcolor{ranktwo}\textit{70.24} \\
        & mcc & \cellcolor{rankone}37.52 & \cellcolor{rankone}34.06 & \cellcolor{ranktwo}55.18 & \cellcolor{rankone}31.88 & \cellcolor{rankthree}35.01 & \cellcolor{ranktwo}40.98 & \cellcolor{rankthree}34.03 & \cellcolor{rankthree}24.55 & 30.63 & \cellcolor{ranktwo}\textit{35.98} \\\bottomrule
        
    \end{tabular}
\caption*{
\centering
\parbox{0.9\linewidth}{\centering
The first, second, and third-best results are highlighted in green, orange, and yellow, respectively. The best value per dataset\\across both tables is shown in \textbf{bold}, while aggregated metrics averaged across all datasets are reported in \textit{italics}.
}
}
\end{table*}

These findings show that model performance depends strongly on both the data split and the characteristics of each dataset. First, strong performance under dissimilarity-based splits does not necessarily imply good generalization to newer molecules, and viceversa. %This discrepancy underscores the risk of relying on a single partitioning strategy for model assessment.

Second, highly imbalanced datasets such as ClinTox or Cardio-30 increase the performance differences among models, particularly for those with limited sensitivity to the minority class, as reflected in their lower MCC values. In contrast, more balanced datasets such as Ames or Tox21 yield higher and more consistent scores across models, reducing the performance gaps.

% The present findings confirm that model effectiveness is profoundly shaped by both the choice of evaluation protocol and the intrinsic properties of the benchmark dataset. Excellence in structural extrapolation —as assessed under the \textit{maxmin} split—does not guarantee comparable proficiency under temporal distribution shifts, and the converse holds equally true, meaning that rankings derived from a single partitioning strategy can be fundamentally misleading. This variability is further complicated by intrinsic dataset characteristics, which introduce additional layers of variability across benchmarks. Highly imbalanced collections such as ClinTox inherently favor methodologies designed to handle minority-class detection —often reflected in the choice of evaluation metrics—, whereas more balanced datasets like Ames yield generally higher and more stable performance across a broader range of architectures. Together, these observations make clear that performance metrics and rankings must be interpreted within their specific experimental context rather than aggregated indiscriminately.

\subsection{Emerging modeling patterns}

Our results reveal that the best-performing approaches, including KPGT, GeoDILI, CD-MVGNN, and 3MTox, incorporate some form of \textit{bond-centered} representations of molecular structure, either explicitly or implicitly. In practice, this is reflected in the use of line graph formulations, edge-aware message passing, or hybrid architectures in which bonds are treated as first-class entities rather than as secondary connections between atoms. From a chemical perspective, this encoding is consistent with the role of bond-dependent properties in toxicity, including molecular reactivity, electronic structure, and metabolic transformations.

By explicitly modeling bond-level information, these approaches allow the network to learn chemical patterns defined by bond types and local bond configurations more directly, rather than inferring them indirectly from atom neighborhoods alone.

Their consistent performance across datasets and partitioning strategies may suggest that bond-centered representations are a promising direction for improving predictive performance while offering more chemically meaningful molecular representations in toxicity modeling.

\section{Conclusions and future directions}
\label{sec:conclusion}

In this work, we conducted a comprehensive scoping review and benchmarking study of graph-based deep learning approaches for molecular toxicity prediction. By integrating diverse and widely used toxicity datasets spanning multiple domains, dataset sizes, and imbalance regimes, we established a unified and publicly available benchmark for standardized, reproducible, and more realistic model evaluation. In particular, the proposed framework emphasizes diverse partitioning strategies to reduce overly optimistic estimates caused by chemical similarity leakage.

By evaluating diverse architectures under consistent and realistic conditions, we identified shared modeling principles that would likely remain obscured by the fragmented, \textit{ad hoc} comparisons commonly found in the state of the art.

Overall, we hope this work provides both a practical evaluation framework and a clearer perspective on the architectural directions that appear most promising for developing more robust, generalizable, and chemically meaningful toxicity prediction models, ultimately supporting their use in real-world drug discovery decision-making pipelines. To this end, we release the full codebase to the scientific community at ~\url{https://gitlab.citius.gal/noel.suarez/benchtox} to enable direct comparison of new models against our publicly available ranking and benchmark results, fostering transparent and reproducible evaluation in this area.

\appendix
% \section{Example Appendix Section}


% %% For citations use: 
% %%       \citet{<label>} ==> Lamport (1994)
% %%       \citep{<label>} ==> (Lamport, 1994)
% %%
% Example citation, See \citet{lamport94}.




\section*{Acknowledgment}

This work has received financial support from the Agencia Estatal de Investigación (Spain) through projects PID2023-149549NB-I00 and PDC2025-166312-I00, as well as from the Xunta de Galicia – Consellería de Educación, Ciencia, Universidades e Formación under the Centro de investigación de Galicia accreditation 2024–2027 (ED431G-2023/04), the Competitive Reference Groups programme (ED431C 2022/19) and the predoctoral grant ED481A-2024-039. This work was also supported by the European Union through the European Regional Development Fund (ERDF).

\section*{Declaration of generative AI and AI-assisted technologies in the manuscript preparation process}

During the preparation of this work, the authors used Perplexity for language refinement and to improve clarity of the text. The authors reviewed and edited the output as needed and take full responsibility for the content of the published article.

\bibliographystyle{elsarticle-num}
\bibliography{references}

\end{document}